\documentclass[journal]{IEEEtran}
\usepackage[noend]{algpseudocode}
\usepackage[ruled,linesnumbered]{algorithm2e}
\usepackage{color}
\usepackage{cite}
\usepackage{graphicx}
\usepackage{subfigure}
\usepackage{float}

\usepackage{amsmath}
\usepackage{amssymb}
\usepackage{multirow}
\usepackage{bm}

\ifCLASSINFOpdf
\else
\fi
\begin{document}
%
\title{Graph-Fraudster: Adversarial Attacks on Graph Neural Network Based Vertical Federated Learning}
%
%
%

\author{Jinyin Chen,
        Guohan Huang,
        Haibin Zheng,
        Shanqing Yu,
        Wenrong Jiang,
        Chen Cui

}

%
%

\markboth{}%
{Shell \MakeLowercase{\textit{et al.}}: Bare Demo of IEEEtran.cls for IEEE Journals}
%



\maketitle

\begin{abstract}
Graph neural network (GNN) has achieved great success on graph representation learning. Challenged by large scale private data collected from user-side, GNN may not be able to reflect the excellent performance, without rich features and complete adjacent relationships. Addressing the problem, vertical federated learning (VFL) is proposed to implement local data protection through training a global model collaboratively. Consequently, for graph-structured data, it is a natural idea to construct a GNN based VFL framework, denoted as GVFL. However, GNN has been proved vulnerable to adversarial attacks. Whether the vulnerability will be brought into the GVFL has not been studied. This is the first study of adversarial attacks on GVFL. A novel adversarial attack method is proposed, named Graph-Fraudster. It generates adversarial perturbations based on the noise-added global node embeddings via the privacy leakage and the gradient of pairwise node. Specifically, first, Graph-Fraudster steals the global node embeddings and sets up a shadow model of the server for the attack generator. Second, noise is added into node embeddings to confuse the shadow model. At last, the gradient of pairwise node is used to generate attacks with the guidance of noise-added node embeddings. Extensive experiments on five benchmark datasets demonstrate that Graph-Fraudster achieves the state-of-the-art attack performance compared with baselines in different GNN based GVFLs. Furthermore, Graph-Fraudster can remain a threat to GVFL even if two possible defense mechanisms are applied. Additionally, some suggestions are put forward for the future work to improve the robustness of GVFL. The code and datasets can be downloaded at https://github.com/hgh0545/Graph-Fraudster.
\end{abstract}


%
\IEEEpeerreviewmaketitle

\section{Introduction}

%
%
%
%
\IEEEPARstart{F}{ederated} learning (FL)~\cite{mcmahan2016federated,yang2019federated}, a privacy-preserving and distributed learning paradigm, establishes machine learning models based on distributed datasets across multiple parties/devices. It aims to protect client's local data and mitigate privacy leakage in the context of legal restrictions, user side privacy protection and commercial competition. Benefiting from the collaborative training of the server model and privacy protection of the local raw data,
FL may be widely applied to various industries, e.g., mobile service~\cite{yang2018applied}, healthcare~\cite{ge2020fedner}, finance~\cite{long2020federated}, face annotation~\cite{suruliandi2021deep, kasthuri2019gabor}. According to the data distribution characteristic, FL can be roughly categorized into horizontal FL (HFL), vertical FL (VFL) and federated transfer learning (FTL). HFL is suitable for the clients sharing datasets of same feature space but different examples, while VFL is designed for the clients sharing datasets of same 
examples but different feature spaces. FTL is proposed for the client's dataset diffs neither in feature space nor in examples. Among the three FLs, most studies are focused on HFL ~\cite{konevcny2016federated, mcmahan2017communication, mohri2019agnostic, yurochkin2019bayesian, mugunthan2020privacyfl}, while the other two still need  to be further explored.

Vertical data distribution is typical in real-world applications, for instance, financial data (e.g., transaction records, income) is often vertically partitioned and owned by different financial institutions (e.g., banks or lending platforms). The banks attempt to avoid lending to users with low credit ratings. Thus, a reliable evaluation agency is necessary to assess the same users among finical parties while sharing different features. To avoid raw data sharing between banks, a VFL framework is a good option for such a practical scenario. As described, some financial data can be constructed as graphs. Consequently, graph neural network (GNN) based VFL, denoted as GVFL, is suitable for this scenario.

\begin{figure}
  \centering 
  \includegraphics[width=3.5 in]{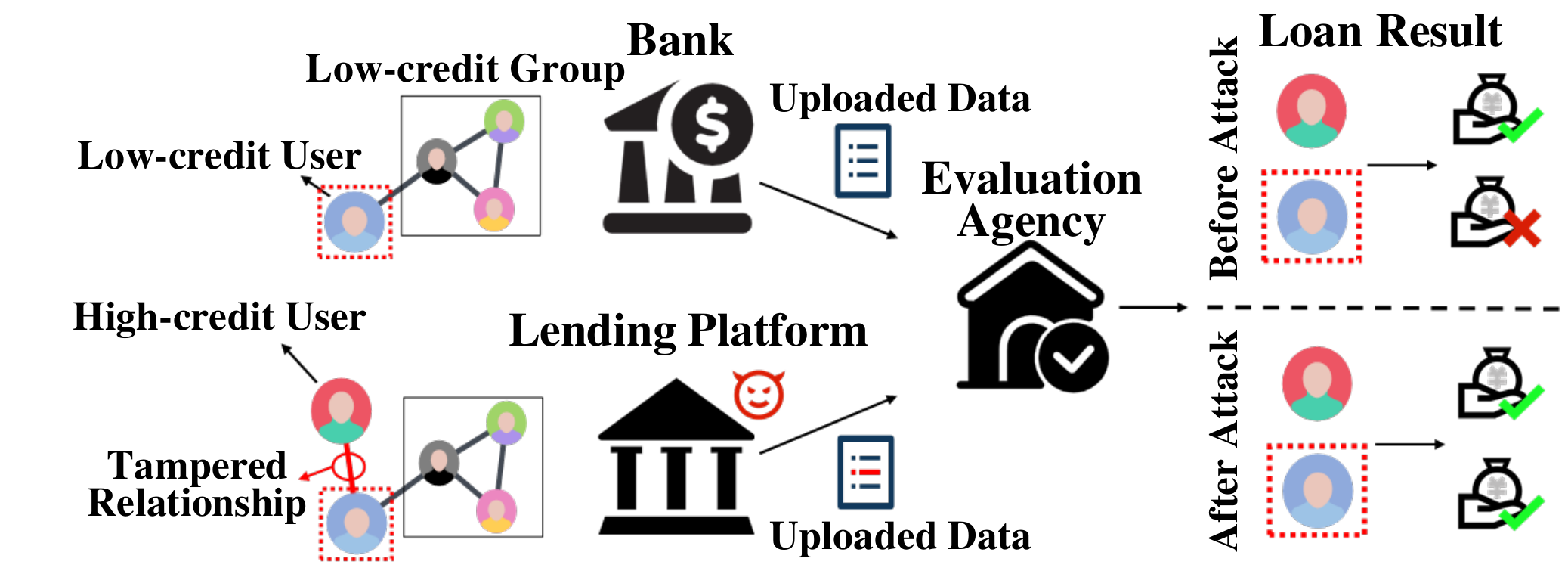}
  \caption{An illustration of the adversarial attack on GVFL. Small perturbations (e.g., fake relationship, extra transaction record) of the graph-structured data could lead to misclassification of the evaluation agency based on GVFL.}
  \label{fig:intro} 
\end{figure}

In such a practical scenario, there could be some potential threats in GVFL. Fig. \ref{fig:intro} shows the threat model of the adversarial attack on GVFL. To be evaluated as high credit rating users, some may hide their true behaviors by adding some extra transaction records. Even worse, some dishonest lending platforms may tamper their data to help the low credit rating users by evading the detection of the evaluation agency. These maliciously crafted operations will cause the bank to issue loans to low-credit users.

We further analyze the possibility of the adversarial attack on GVFL, and find out that the threats could come from the innate deficiency in GVFL. Specifically, each participant in GVFL uploads the intermediate information, i.e., the node embeddings extracted by the local GNN model instead of the raw data. The adversarial perturbations (e.g., fake relationship, extra transaction record) on the local raw data will influence the updated node embeddings, thus posing a threat to the server model by making a wrong decision. Besides, privacy leakage of embeddings of GVFL will provide conditions for adversarial attacks, e.g., the leakage of the embeddings will help the malicious client to establish a precise shadow model of the server. These threats of adversarial attacks on GVFL have not been studied yet. Motivated by the practical application of GVFL and its potential vulnerability towards adversarial attacks, a novel attack method is proposed in this paper as the first work of security research of GVFL.

The contributions of our work are summarized as follows:

\begin{itemize}
\item To the best of our knowledge, this is the first work of proposing and formulation the adversarial attack on GVFL, by revealing its vulnerability towards a crisis of distrust in practical applications. Due to privacy leakage and data bias of the local model, GVFL unintentionally provides the conditions for successful adversarial attacks. 
\item A novel adversarial attack method is proposed against GVFL, denoted as Graph-Fraudster. Aiming at confusing the server model in GVFL, Graph-Fraudster generates adversarial perturbations based on the noise-added global node embeddings via GVFL’s privacy leakage and the gradient of pairwise node.
\item Extensive experiments are conducted on five real-world graph-structured datasets for node classification with different GNN structured GVFLs. Graph-Fraudster is validated that it performs significantly better than three advanced adversarial attacks transferable to GVFL from centralized framework on seven metrics, achieving the state-of-the-art performance.
\item Furthermore, we propose two possible defenses against the attack. Graph-Fraudster can still pose a threat to the defensive GVFL. Additionally, according to the observations of the experiments, some suggestions are provided to improve the robustness of GVFL, including privacy preserving, bias removal and defense/detection deployment. The code and datasets are available in https://github.com/hgh0545/Graph-Fraudster.
\end{itemize}

The rest of the paper is organized as follows. Related works on GNN based federated learning, inference attacks on federated learning and adversarial attacks on GNN models are reviewed in Section~\ref{rw}. Section~\ref{method} defines the adversarial attack on GNN model and GVFL, sets up threat model, and then introduces Graph-Fraudster in details. In Section~\ref{experiment}, extensive experiments are presented to demonstrate the performance of Graph-Fraudster. At last, the paper is concluded, and the future work is enumerated in Section~\ref{conclusion}. 

\section{Related Work}

In line with the focus of the work, we briefly summarize the existing works of GNN based federated learning,  inference  attacks  on  federated learning and adversarial attacks on GNN.
\label{rw}
\subsection{GNN based Federated Learning}
Though FL has shown the superiority in various domains, it has not been widely applied to graph-structured data. According to the investigation, most of the research focus on horizontal scenarios. For instance, considering data privacy preservation in non-independent identically and distribution (Non-IID), Zheng et al.~\cite{zheng2020asfgnn} used clients to implement message communication, and leveraged Bayesian optimization for hyper-parameters fine-tuning. They further proposed a novel learning paradigm, named ASFGNN. Inspired by meta learning, GraphFL~\cite{wang2020graphfl} is proposed to address the node classification under Non-IID. These proposed FLs are supported by machine learning models designed for graph mining. In the aspect of GNN as the local model for FL, He et al.~\cite{he2021fedgraphnn} presented FedGraphNN, an open-source FL system based on GNN models. But it is not suitable for VFL. As for GNN based vertical federated learning, the research is still in its fancy. Zhou et al.~\cite{zhou2020privacy} gave the first vertical learning paradigm for privacy-preserving GNN models for node classification, by splitting the graph into two parts for different clients. Secure muti-party computation is adopted as well to ensure data privacy and efficiency. To ensure privacy, Ni et al.~\cite{ni2021vertical} adopted additively homomorphic encryption, and proposed a federated GCN learning paradigm for the privacy-preserving node classification task, named FedVGCN. Besides the framework study of GNN based FL, it is also applied to real-world applications. For instance, Wu et al. ~\cite{wu2021fedgnn} proposed a GNN structured FL for recommendation to accomplish server model training and user-side privacy. 

\subsection{Inference Attack on Federated Learning}
Since the inference attack on GNN based federated learning has not been studied yet, we summarize some inference attacks on federated learning. 
Numerous attacks on HFL have been proposed. For instance, in~\cite{hitaj2017deep}, generative adversarial network is used for inference attack on HFL for the first time, which generates the same distribution as the training data. Nasr et al.~\cite{nasr2019comprehensive} exploited the vulnerability of the stochastic gradient descent (SGD), and designed a membership attack in white-box setting. Besides, deep leakage from gradients (DLG)~\cite{zhu2020deep} uses public shared gradients to obtain the local training data without extra knowledge of the data. To sum up, most of them greatly rely on the gradient which exchanged during training process. In vertical scenarios, Luo et al.~\cite{luo2020feature} formulated the problem of feature inference attack in model's inference process for the first time, and they proposed two attack methods on the logistic regression (LR) and decision tree (DT) models, named equality solving attack (ESA) and path restriction attack (PRA), respectively. Further, they proposed a general attack method named generative regression network attack to handle more complex models. Zhang et al.~\cite{zhang2021privacy} devoted to gathering the man-made poison data and the corresponding intermediate outputs in the model's training process. Then they built the training set for the decoder to recover other private data of the victim. Besides, Li et al.~\cite{li2021label} explored whether the labels can be uncovered by sharing gradient in the VFL setting, and proposed a label inference attack method based on the gradient norm.

\subsection{Adversarial Attacks on GNN Models}
Existing technologies may have multiple vulnerabilities that enable hackers or criminals to manipulate data for malicious purposes~\cite{iwendi2021sustainable}, and the adversarial examples may jeopardize the operation of the reality systems such as the Internet of Things~\cite{ahmed2021generative}. Extensive studies have proven that GNN models are vulnerable to imperceptible adversarial perturbations that affect the performance of downstream applications. For node classification, Zügner et al.~\cite{zugner2018adversarial} proposed the first adversarial attack on GNN models, denoted as NETTACK, by generating attacks iteratively according to score functions. Reinforcement learning is applied to generate perturbations in~\cite{dai2018adversarial} as well. Aiming at fooling GNN models by attacking embeddings of the target node, Chen et al.~\cite{chen2018fast} proposed a fast gradient attack method (FGA) via the maximal absolute edge gradient. Analogously, IG-FGSM and IG-JSMA are introduced in~\cite{wu2019adversarial}. To deal with the large-scale graph, Li et al.~\cite{li2021adversarial} proposed a multi-stage attack framework SGA, which relies on a much smaller subgraph centered at the target node. Another modification strategy is proposed by inserting fake nodes~\cite{wang2018attack} instead of adding/deleting edges in the adversarial examples. Considering a black-box scenario, GF-Attack~\cite{chang2020restricted} is constructed by the graph filter and feature matrix without accessing any knowledge of the target classifiers. For graph classification, Wu et al.~\cite{ma2019attacking} used rewiring to hide the adversarial edges, which preserves some important properties of the graph such as the number of nodes, edges and total degrees of the graph. For community detection, genetic algorithm is adopted in Q-Attack~\cite{chen2019ga} to fail the detection method. To sum up, the existing graph adversarial attacks mainly degrade the performance of GNN models in various tasks by adding or deleting key edges.

\section{Methodology}
\label{method}
First, we formalize the problem of adversarial attack on GVFL, and give its definition. Then, the threat model of adversarial attack on GVFL is elaborated. Afterward, we introduce the Graph-Fraudster in aspect of framework, implementation details and algorithm rationality. For convenience, the definitions of symbols used in this paper are listed in the TABLE~\ref{Symbol}.
\begin{table}[!ht]
    \centering
    \caption{The definitions of symbols.}
    \label{Symbol}
    \begin{tabular}{c|r}
    \hline \hline
    Symbol&Definition\\ \hline
    $G=(V,E)$ &the original graph with sets of nodes and edges \\
    $A,\hat{A}$ & the adjacency matrix / adversarial adjacency matrix of $G$\\
    $X,\hat{X}$ & the feature matrix / adversarial feature matrix of nodes\\
    ${f}_{\theta }(\cdot)$ & the GNN model with parameter $\theta$\\
    $K$&the number of participants\\
    $h_{m}$ & the node embeddings of the malicious participant \\
    $h_{global}$ & the global node embeddings\\
    $V_{L}$ & the set of nodes with labels\\
    $T$ & the set of target nodes\\
    $\rho$ & the activation function\\
    $Y,Y^{'}$ & the real / predicted label list\\
    $|F|$ & the number of class for nodes in the $G$\\
    $N$ & the number of nodes in the graph $G$\\
    $P$ & the probabilities with the real global node embeddings\\
    $\Tilde{P}$ & the probabilities with the fake global node embeddings\\
    $d$ & the dimensions of local node embeddings\\
    $h_{fake}$ & the fake global node embeddings\\
    $\mathcal{S}(\cdot)$ & the server model in GVFL\\
    $\Tilde{\mathcal{S}}(\cdot)$ & the shadow server model\\
    $C^{0}_{d}, R^{0}_{d}$& the weights and bias of the simplified $\mathcal{S}(\cdot)$\\
    $B_{S}, B_{\tilde{S}}$& the boundary of $\mathcal{S}$ and $\Tilde{\mathcal{S}}$\\
    \hline \hline
    \end{tabular}
\end{table}
\begin{figure}
  \centering 
  \includegraphics[width=2.3 in]{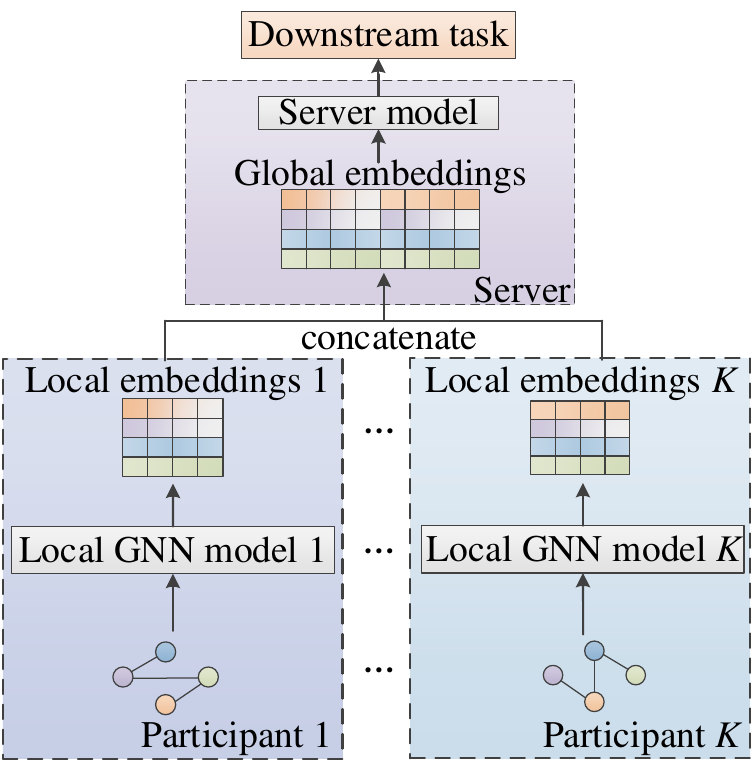}
  \caption{Forward propagation of GVFL. The server model concatenates the local embeddings as global embeddings, and completes the downstream task.}
  \label{fig:VFL} 
\end{figure} 
\subsection{Problem Definition}
\textbf{Definition 1: Adversarial attack on GNN model.} For node classification, the attacker is aiming to mislead the target GNN model to output the expected label of the adversarial example, which is carefully crafted with imperceptible perturbations (e.g., rewiring edges, fake nodes) on the benign one. As expected, fed by the adversarial example, the GNN model will extract low quality node embeddings, leading to the wrong prediction. Specifically, $G=(V, E)$ represents a graph and $f_\theta(\cdot)$ represents a GNN model. Denote $A$ and $X$ as the adjacency matrix of $G$ and node features, respectively. The adversarial attack on GNN model can be formalized as,
\begin{equation}
	\begin{split}
    & \max \sum\limits_{t\in T}{{{L}_{atk}}(softmax({{f}_{{{\theta }^{*}}}}(\hat{A},\hat{X},{{v}_{t}})),{{y}_{t}})} \\
    & s.t.\quad {{\theta }^{*}}=\underset{\theta }{\mathop{\arg \min }}\sum\limits_{{{v}_{l}}\in {{V}_{L}}}{{{L}_{train}}({{f}_{\theta }}(A,X,{{v}_{l}}),{{y}_{l}})}
	\end{split}\label{1}
\end{equation}
where $\hat{A}$ represents the perturbed adjacency, $\hat{X}$ expresses the perturbed node features. ${v}_{t}$ represents the target node and ${y}_{t}$ is the ground truth of ${v}_{t}$. The parameters of GNN model $\theta$ are trained with labeled node sets ${V}_{L}$, resulting ${\theta }^{*}$. Note that, the attacker can manipulate edges and node features respectively, or both. The research shows that modifying edges is more effective than modifying node features~\cite{wu2019adversarial}. Thus, only adding/deleting edges is considered as perturbations in this work.

\textbf{Definition 2: Adversarial attack on GVFL.} In GVFL, each client trains a local GNN model with its private data, and updates the embedding for the aggregation of the server. The attacker's goal is to disrupt the local embeddings, causing the server model to produce wrong predictions.  Fig. \ref{fig:VFL} is a schematic diagram of the forward propagation for GVFL. 

In this paper, concatenating the local node embeddings is used as the combination strategy:
\begin{equation}
    h_{global}=h_{1}||h_{2}||\dots||h_{K}
\end{equation}
where $K$ denotes the number of participants, and $||$ is the concatenating operator. 

\begin{figure*}
  \centering 
  \includegraphics[width=7in]{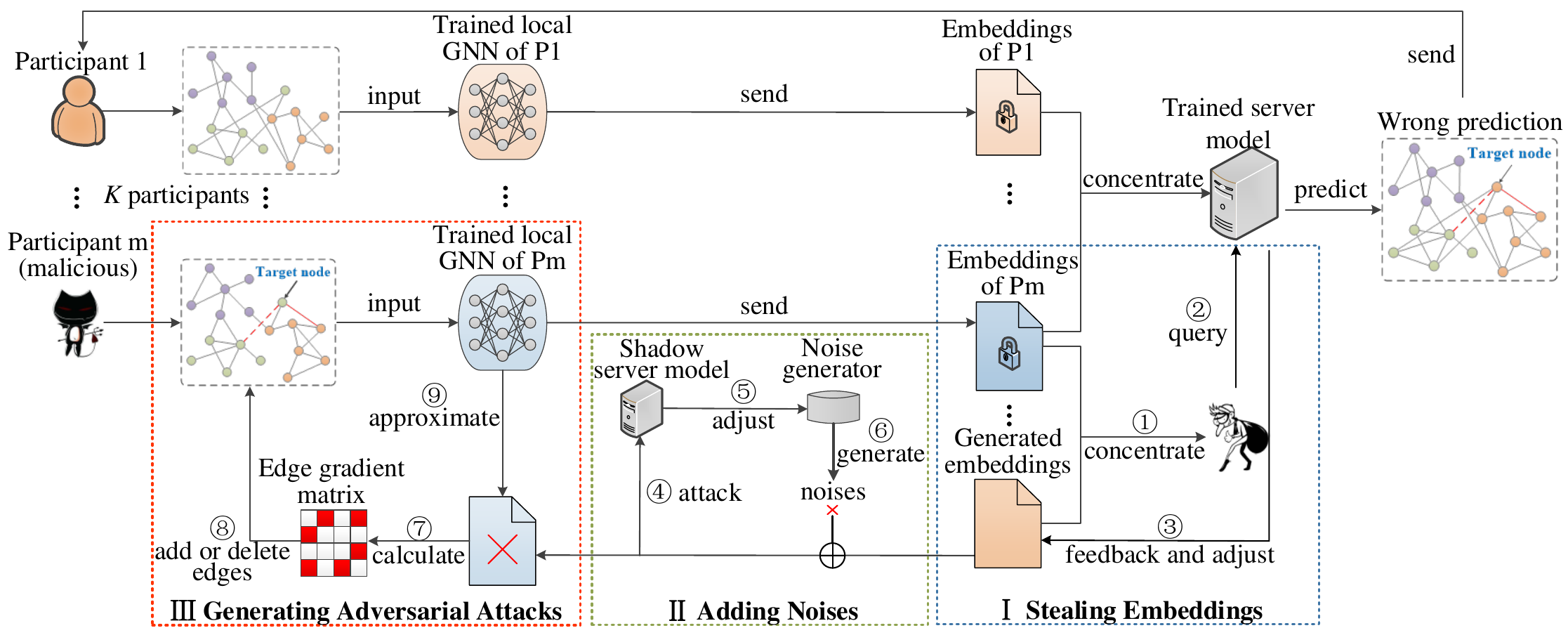}
  \caption{The framework of Graph-Fraudster. One participant is randomly selected as the malicious participant from all. The adversarial attack is conducted through three stages: (1) stealing embeddings: a global node embeddings is generated, which adjusted by querying result of trained server model; (2) adding noises: noises are added into the generated embeddings to attack the shadow model of the server; (3) generating adversarial attacks: guided by noise-added node embeddings of malicious participant, adversarial attacks are generated by using gradient of pairwise node. As a result, the wrong prediction of the target node will be produced by the trained server model.}
  \label{fig:framework} 
\end{figure*}

For a node classification task, the server model $\mathcal{S}$ is a classifier to make prediction by:
\begin{equation}
    Y^{'}=softmax(W_{l}\cdot\rho(\dots\rho(W_{0}\cdot h_{global})))
\end{equation}
where $\{W_{0} \dots W_{l}\}$ are weight matrices of the classifier. $\rho(\cdot)$ represents activation function such as ReLU.

Further, the GVFL uses the cross-entropy error over all labeled examples, which can be learned by gradient descent:
\begin{equation}
    L_{train}=-\sum_{l=1}^{|V_L|}\sum_{n=1}^{|F|}Y_{ln}\ln(Y^{'}_{ln})
\end{equation}
where $V_{L}$ is the set of nodes with labels, $Y_{ln}$ is the ground truth, and $|F|$ is the number of node classes in the graph $G$.

Then, the adversarial attack on GVFL is defined as:
\begin{equation}
    \footnotesize
    \begin{split}
   \hat{Y}_{t}^{'}=softmax &\left(W_{l}\cdot\rho(\dots\rho(W^{v_{t}}_{0}\cdot (h^{v_{t}}_{1}||h^{v_{t}}_{2}||\dots||\hat{h}^{v_{t}}_{m}||h^{v_{t}}_{K})))\right)\\
    &s.t. \quad \hat{h}^{v_{t}}_{m} = f_{{\theta }^{*}}(\hat{A},\hat{X},v_{t})
    \end{split}
\end{equation}
where $\hat{h}^{v_{t}}_{m}$ is the perturbed node embeddings of the target node $v_{t}$ which is uploaded by a malicious participant. It means a malicious participant can add perturbations into the graph by manipulating $A$ or $X$. Then, the local GNN model may be confused, which will produce low quality or even targeted node embeddings and upload the perturbed node embeddings to the server model. Therefore, the server model will make wrong decision. 

The target loss function $L_{atk}$ of target node $v_{t}$ can be defined as:
\begin{equation}
    L_{atk}=-\sum_{n=1}^{|F|}Y_{tn}\ln(Y^{'}_{tn})
\end{equation}
which measures the difference between the prediction and the ground truth. It is easy to find the prediction of the model is worse with larger value of $L_{atk}$.

\subsection{Threat Model}
\label{threat}

\textbf{Scenario.} Taking more general scenarios into consideration, GVFL is applied to multiple participants training a central server collaboratively for node classification. When the server model is well-trained, the prediction of the server model will be shared by all participants. Besides, the server model is semi-honest, i.e., the server model serves to be queried by the malicious participant without revealing the inner parameters.

\textbf{Knowledge of the malicious participant.} Assume that the malicious participant knows the structure of the server, and only can access to its own data. In other word, the malicious participant can attack the server model by manipulating its own data merely. In addition, the trained server model will be served as an API for the malicious participant, and it only returns probability for each query. Thus, more details, such as the parameters of the server model and the gradient information, are unable to be obtained by the malicious participant.

\textbf{The goal of the malicious participant.} For the malicious participant, the goal is to attack the server model by uploading perturbed embeddings. So that the other benign participants, which rely on the server model, receive the unexpected prediction for the target example.

\begin{figure}
  \centering 
  \includegraphics[width=3.5 in]{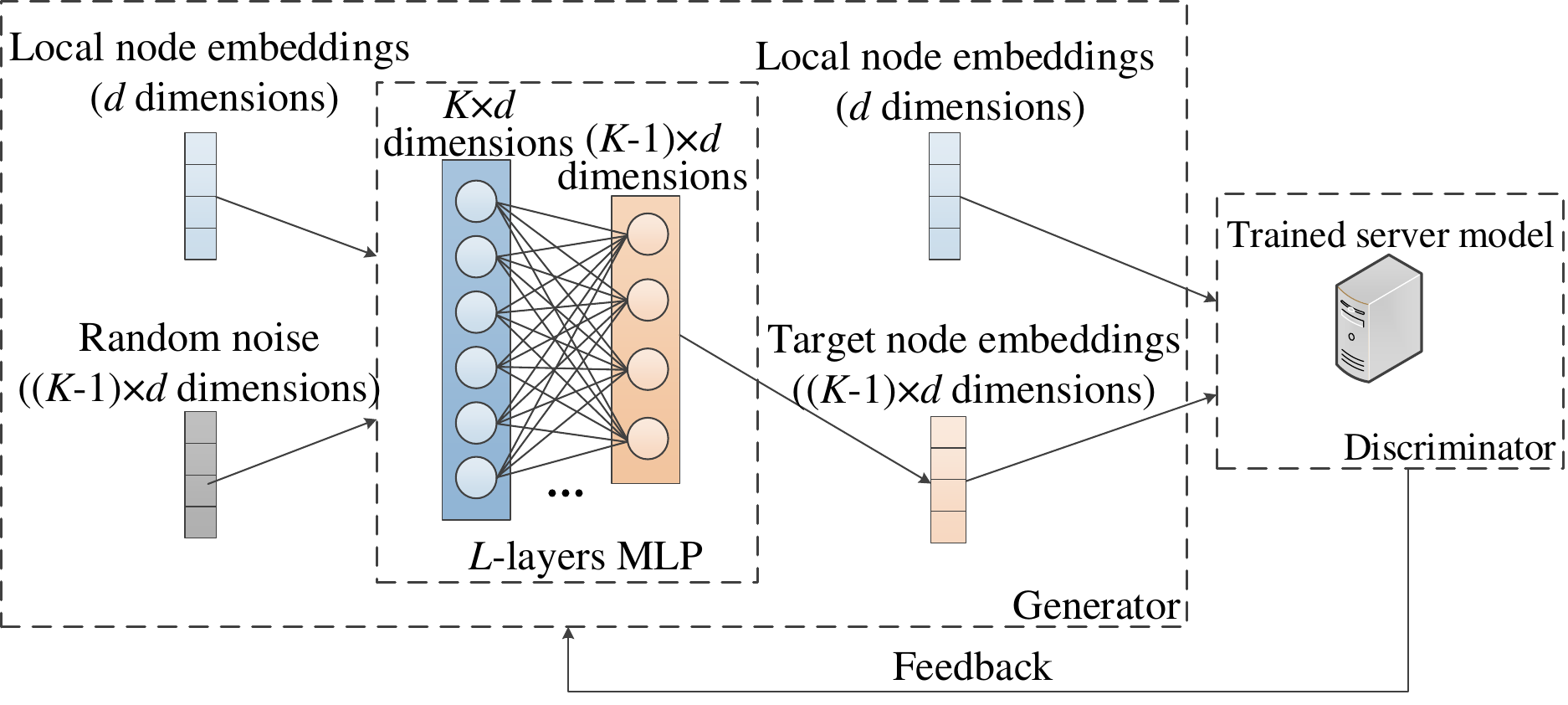}
  \caption{Stealing node embeddings via GRN.}
  \label{fig:GRN} 
\end{figure}

\subsection{Graph-Fraudster}
\label{Graph-Fraudster Method}
As described in Section~\ref{threat}, the attacker can manipulate the own edges or node features in the testing to fool the global GVFL. During the aggregate process, modifying the edges affects all dimensions of features belonging to the target node, while modifying features only affects a few features. Thus, Graph-Fraudster is designed to attack the server model by manipulating edges of the local graph. It is conducted through three stages: (1) stealing node embeddings of other participants; (2) adding noises into embeddings via constructing the shadow server model; (3) generating adversarial attacks to mislead the global GVFL. The framework of Graph-Fraudster is shown as Fig. \ref{fig:framework}. The details and rationality of Graph-Fraudster is described in this subsection.

\subsubsection{Embeddings Stealing Strategy} Similar to~\cite{luo2020feature}, Graph-Fraudster uses generative regression network (GRN) to steal embeddings uploaded by other participants. First, as shown in Fig. \ref{fig:GRN}, a $(K-1)\times d$ random noise vector is generated, where $K$ is the number of participants in GVFL, and $d$ is dimensions of local node embeddings. Then, a $L$-layers multilayer perceptron (MLP) is used to generate the target node embeddings, i.e., the embeddings uploaded by other participants. In this paper, a three-layer MLP with ReLU is applied as the generator. The dimension of an input is $K\times d$, the dimension of an output is $(K-1)\times d$, and the hidden units are 512 and 256, respectively. Then, the local node embeddings of the malicious participant and the target node embeddings are concatenated, represented as $h_{fake}$, as input of the discriminator. The classification probabilities of an adversarial example can be obtained by querying the server model. Note that the parameters of the trained server model are fixed. The target of stealing privacy can be converted to a regression problem. Therefore, mean square error (MSE) is adopted for the training process, which can be formulated as:
\begin{equation}
    L_{GRN}=\frac{1}{N\times |F|}\sum_{i=1}^{N}\sum_{j=1}^{|F|}(\Tilde{P}_{i,j}-P_{i,j})^{2}\label{lgrn}
\end{equation}
where $N$ is the number of nodes, $\Tilde{P}=\mathcal{S}(h_{fake})$ is the probabilities returned when node embeddings of an adversarial example are fed. $P$ is the probabilities returned at the end of GVFL's training process. 

To recognize the GRN in details, we simplify the model structure of GRN with a single-layer generator model and a discriminator model $S$ with fixed parameters. Thus, for the real node embeddings ${{h}_{real}}$, its ground-truth prediction output of the discriminator model can be represented as
    \begin{equation}
    P=\mathcal{S}({{h}_{real}})= softmax (C_{d}^{0}{{h}_{real}}+R_{d}^{0})
    \end{equation}
    
Due to the parameters of $\mathcal{S}$ are fixed, $C_{d}^{0}$ is used to represent the weights of $\mathcal{S}$. $R_{d}^{0}$ is bias of $\mathcal{S}$. 
For a fake node embeddings $h_{fake}$, the output of $\mathcal{S}$ is
  \begin{equation}
      \begin{split}
           {{P}_{1}}&=\mathcal{S}({{h}_{fake}})\\
           &=\mathcal{S}(g({{n}_{in}})||{{h}_{m}})\\
          &=softmax (((({{n}_{in}}||{{h}_{m}}){{w}_{0}}+{{b}_{0}})||{{h}_{m}})C_{d}^{0}+R_{d}^{0}) \\ 
         & s.t.\quad g({{n}_{in}})=({{n}_{in}}||{{h}_{m}}){{w}_{0}}+{{b}_{0}} \\
      \end{split}
  \end{equation}
where ${n}_{in}$ is a $(K-1)\times d$ dimensions noise. ${h}_{m}$ is local node embeddings of the malicious participant, $g(\cdot)$ is the generator, ${{w}_{0}}$ is the weights of the generator model and ${{b}_{0}}$ is bias. GRN back-propagates the loss and updates the parameters of the MLP model.

\subsubsection{Noise Addition via Stolen Embeddings}
In order to degrade the performance of the server model, it's significant to produce the adversarial embeddings crossing the decision boundary to mislead the server model.  An effective and efficient method is using the gradient information of the server model to produce adversarial examples, which has been widely used to construct attack on GNN. However, in GVFL, the attackers cannot get gradient information but only the returned probabilities. Therefore, it is necessary to establish a shadow model in local to simulate the server model. By stealing the global node embeddings, the shadow model $\Tilde{\mathcal{S}}$ can be successfully established based on the necessary knowledge, i.e., training data $h_{fake}$, model structure and target probabilities $P$. According to the setting in Section~\ref{threat}, the model structure and $P$ are known. To the end, the shadow model will output the similar probability $P$ to the server's when fed by the same examples. MSE is applied as the target as follows:
\begin{equation}
    L_{shadow}=\frac{1}{N\times |F|}\sum_{i=1}^{N}\sum_{j=1}^{|F|}(\Tilde{\mathcal{S}}(h_{fake})-P_{i,j})^{2} \label{lshadow}
\end{equation}
where $\Tilde{\mathcal{S}}(h_{fake})$ is the output of $\Tilde{\mathcal{S}}$. 

Then, fast gradient sign method (FGSM)~\cite{goodfellow2014explaining} is adopted as noise generator by using the gradient information of the shadow model $\Tilde{\mathcal{S}}$.

For a target node $v_{t}$, its noise-added embeddings $\overline{h}^{v_{t}}_{fake}$ could be represented as:
\begin{equation}
    \overline{h}^{v_{t}}_{fake} = h^{v_{t}}_{fake}+\epsilon \cdot sign(\frac{\partial L_{atk}}{\partial h^{v_{t}}_{fake}} )\label{noise}
\end{equation}
where $sign(\cdot)$ is gradient's direction, $\epsilon \in [0,1]$ is noise scale. 

In particular, the reason for using MSE as the target to establish the shadow model is as follows.
Suppose using a mapping equation to represent the server model:
    \begin{equation}
        {{\hat{k}}_{S}}(h)=\underset{k}{\mathop{\arg \max }}\,\mathcal{S}(h)
    \end{equation}
where $h$ is the node embeddings and $k$ is $k$-th class of $\mathcal{S}$. The classification boundary function can be express as:
    \begin{equation}
        \mathcal{S}_{k}(h)-\hat{k}_{\mathcal{S}}(h)=0
    \end{equation}
    
By applying the first-order Taylor expansion, the approximate classification boundary function of $\mathcal{S}$ is denoted as $\tilde{\mathcal{S}}$:
    \begin{equation}
        \begin{split}
            & {{B}_{\mathcal{S}}}:{{\mathcal{S}}_{k}}({{h}_{real}})+{{h}_{real}}\nabla S_{k}({{h}_{real}})\\
            &-{{{\hat{k}}}_{\mathcal{S}}}({{h}_{real}})-{{h}_{real}}\nabla {{{\hat{k}}}_{\mathcal{S}}}({{h}_{real}})=0 \\ 
        \end{split}
    \end{equation}
    \begin{equation}
        \begin{split}
            & {{B}_{{\tilde{\mathcal{S}}}}}:{{{\tilde{\mathcal{S}}}}_{k}}({{h}_{fake}})+{{h}_{fake}}\nabla \tilde{\mathcal{S}}_{k}({{h}_{fake}})\\
            &-{{{\hat{k}}}_{{\tilde{\mathcal{S}}}}}({{h}_{fake}})-{{h}_{fake}}\nabla {{{\hat{k}}}_{{\tilde{\mathcal{S}}}}}({{h}_{fake}})=0 \\ 
        \end{split}
    \end{equation}
    
The MSE of the shadow model is aiming to approximate the classification boundary of $\tilde{\mathcal{S}}$ and $\mathcal{S}$. Thus, the target can be converted to:
    \begin{equation}
        \begin{split}
              & \underset{{\tilde{\mathcal{S}}}}{\mathop{\arg \min }}\,MSE(\tilde{\mathcal{S}}({{h}_{fake}}),\mathcal{S}({{h}_{real}})) \\ 
 & \Leftrightarrow \underset{{\tilde{\mathcal{S}}}}{\mathop{\arg \min }}\,|{{B}_{{\tilde{\mathcal{S}}}}}-{{B}_{\mathcal{S}}}| \\ 
 & \Leftrightarrow \underset{{\tilde{\mathcal{S}}}}{\mathop{\arg \min }}\,|\tilde{\mathcal{S}}({{h}_{fake}})-\mathcal{S}({{h}_{real}})|\\
 &+|({{h}_{fake}}-{{h}_{real}})(\nabla \tilde{\mathcal{S}}({{h}_{fake}})-\nabla \mathcal{S}({{h}_{real}}))| \\ 
        \end{split}
    \end{equation}
    
Then, assuming $\nabla \tilde{\mathcal{S}}({{h}_{fake}})\approx \nabla \mathcal{S}({{h}_{real}})$, and the noise generated by FGSM against the shadow model can also attack the server model in GVFL successfully since the approximate direction of the gradient is obtained.

\subsubsection{Adversarial Attack via Noise-added Embeddings} 
Based on the above steps, the noise-added embeddings $\overline{h}^{v_{t}}_{fake}$ are obtained.  Next, the part of the noise-added embeddings $\overline{h}^{v_{t}}_{m}$ that belongs to the malicious participant is extracted from $\overline{h}^{v_{t}}_{fake}$. Assuming that $\bar{h}_{fake}^{{{v}_{t}}}$ can confuse the shadow model successfully, the rest of the work is modifying some suitable edges in the original graph to make the node embeddings approximate to $\bar{h}_{m}^{{{v}_{t}}}$. Thus,  Graph-Fraudster devotes to making the target node's embeddings and $\overline{h}^{v_{t}}_{m}$ more similar by adding perturbations into $A$.

\begin{equation}
    \underset{\hat{A}}{\mathop{\arg \min }}\quad MSE(f_{{\theta}^{*}}(\hat{A},X,v_{t}),\overline{h}^{v_{t}}_{m})\label{ad_target}
\end{equation}

where $\parallel \hat{A} -A \parallel _{0}$ is twice the budget $\Delta$, i.e., $\parallel \hat{A} -A \parallel _{0} \leq 2\Delta$ due to the symmetry of the adjacency matrix.

To speed up the process of searching the suitable adversarial edges, the gradient of pairwise node is applied, followed the work of Chen et al.~\cite{chen2018fast}. The edge gradient matrix is computed as follows:
\begin{equation}
    \begin{split}
     &g_{u,v}=\frac{\partial L_{t}}{\partial A_{u,v}}\\
    s.t. \quad L_{t} = \frac{1}{d}\sum_{i=1}^{d}(&[f_{{\theta}^{*}}(\hat{A},X,v_{t})]_{i}- [\overline{h}^{v_{t}}_{fake}]_{i})^{2}
    \end{split} \label{grad}
\end{equation}
where $(u,v)$ is any pairwise node in the graph $G$, $[\cdot]_{i}$ denotes the $i$-th element of node embeddings and $d$ is the dimensions of local node embeddings. Considering the symmetry of the adjacency matrix of the undirected graphs, the gradient matrix is symmetric as:
\begin{equation}
    g_{sym} = \frac{g + g^{T}}{2} \label{symgrad}
\end{equation}
where $\cdot^{T}$ is the transpose symbol. The adversarial edge $e_{adv}$ is selected by:
\begin{equation}
    e_{adv}=(u,v) \xleftarrow{} \underset{u,v}{\mathop{\arg \max }} - g_{sym} \label{ads}
\end{equation}

To minimize the difference between the local GNN model's output and the noise-added embeddings rather than maximizing $L_{atk}$, the gradient value should be inverted. At last, the adversarial edges will be added into clean adjacency iteratively within budget $\Delta$:
\begin{equation}
    \hat{A}_{u,v} = -A_{u,v}+1 \label{adadd}
\end{equation}
where $(u,v)$ comes from $e_{adv}$ and $\hat{A}$ is symmetrical.

\subsection{Algorithm}
The pseudo-code for Graph-Fraudster is given in Algorithm~\ref{alg1}.

\begin{algorithm}
\label{alg1}
\small
\caption{Graph-Fraudster}
\KwIn{Original adjacency $A$, node features $X$, target node $v_{t}$, attack budget $\Delta$, dimensions of local node embeddings $d$, number of participants $K$.}
\KwOut{The adversarial network $\hat{A}$.}
 Train the GVFL to obtain the server model $\mathcal{S}$ and the local GNN models $f_{{\theta }^{*}}(\cdot)$.\\
 Generate a $(K-1)\times d$ dimensions noise randomly.\\
 Concatenate generated noise and the local node embeddings of malicious participant $h_{m}$ as input $h_{in}$.\\
 \For{$t=1$ to $T_{GRN}$}
 {
    $h_{target} \xleftarrow{} MLP(h_{in})$\\
    Concatenate $h_{target}$ and $h_{m}$ as $h_{fake}$ for the server model $\mathcal{S}$.\\
    Minimize the MSE in Equation~\ref{lgrn}.\\
 }
 Save $h_{fake}$.\\
 Initialize the parameter of the shallow server model $\Tilde{\mathcal{S}}$.\\
\For{$t=1$ to $T_{shadow}$}
{
    Minimize the MSE in Equation~\ref{lshadow} with $h_{fake}$.
}
Generate the noises and add them into $h_{fake}$ as Equation~\ref{noise} and return $\overline{h}^{v_{t}}_{fake}$ for target node $v_{t}$.\\
Initialize $\hat{A}^{0}=A$.\\
\For{$i=1$ to $\Delta$}
{
Calculate the symmetrical edge gradient matrix as Equation~\ref{grad} and ~\ref{symgrad}.\\
Select adversarial edge $e_{adv}=(u,v)$ by Equation~\ref{ads}.\\
Add adversarial edge $e_{adv}=(u,v)$ into $\hat{A}^{i-1}$ as  Equation~\ref{adadd} and obtain $\hat{A}^{i}$.\\
}

\textbf{Return} the adversarial adjacency matrix $\hat{A}$.\\
\end{algorithm}

\subsection{Complexity Analysis}
For generation of adversarial attacks, the calculation time of Graph-Fraudster is divided into four parts, including GRN’s training time $T_{GRN}$, shadow model’s training time $T_{shadow}$, noise adding time and adversarial edges selecting time. Thus, the time complexity of Graph-Fraudster is:           
\begin{equation}
    {\mathcal O}({{T}_{GRN}})+{\mathcal O}({{T}_{shadow}})+{\mathcal O}(1)+{\mathcal O}(\Delta)\sim {\mathcal O}(N)
    \end{equation}
where $\Delta$ is the attack budget. ${\mathcal O}({{T}_{GRN}})$ and ${\mathcal O}({{T}_{shadow}})$ are the complexity of the training time of GRN and the shadow, depending on maximum training epochs ${T}_{GRN}$ and ${{T}_{shadow}}$, respectively. ${\mathcal O}(1)$ indicates the complexity of noise adding time because FGSM is a one-step noise generator. ${\mathcal O}(\Delta)$ is the complexity of adversarial edges selecting time, which controlled by the attack budget $\Delta$. ${\mathcal O}(N)$ indicates the complexity of Graph-Fraudster is linear, according to the total number of the above steps $N$.

The parameters of Graph-Fraudster include GRN’s parameters, shadow model’s parameters. Therefore, the space complexity is:
  \begin{equation}
  \begin{aligned}
   &{\mathcal O}(K\times d\times {{n}_{0}}+{{n}_{0}}\times {{n}_{1}}+\ldots +\\&(K-1)\times d\times {{n}_{L-1}})+{\mathcal O}(\tilde{S})
   \sim {\mathcal O}({{M}^{2}})
   \end{aligned}
    \end{equation}
where $K$ is the number of participants, $d$ is the dimensions of local node embeddings, $[n_{0}, \ldots,n_{L-1}]$ is the number of the units in GRN models and $\tilde{S}$ is the shadow model. ${\mathcal O}({{M}^{2}})$ indicates that the space complexity depends on the memory occupied by the model’s weight matrix, whose is squared-level.

\section{Experiments}
\label{experiment}
In order to comprehensively evaluate the performance of Graph-Fraudster, both dual-participants based GVFL and multi-participant based GVFL are testified in aspect of attack performance, two possible defense strategies against GVFL and parameter sensitivity analysis.

\subsection{Datasets}
Five graph-structured datasets are used to evaluate the performance of the proposed method, including Cora~\cite{mccallum2000automating}, Cora\_ML~\cite{mccallum2000automating}, Citeseer~\cite{mccallum2000automating}, Pol.Blogs~\cite{adamic2005political} and  Pubmed~\cite{sen2008collective}. Their basic statistics are summarized in TABLE~\ref{Datasets}, and the specific information is the same as the reference source.

In GVFL, each dataset will be split for different clients. Assuming there are no overlap edges for each client, then with more clients participant in, the more isolated nodes appear in clients' graph data. It will be not conducive to collaboratively train the server model. 
Consequently, two splitting strategies are adopted for different number of clients. Specifically, for dual-participants, both edges and node features of the dataset are divided into two parts on average, in which a small part of isolated nodes will appear. For multi-participant, node features are divided averagely to each participant, and the complete topology of the graph is retained. Due to the randomness of the segmentation, the dataset is divided and tested for 10 times. The average accuracy of the trained server model will be recorded.

\begin{table}[h]
    \centering
    \caption{The basic statistics of five network datasets}
    \label{Datasets}
    \setlength{\tabcolsep}{1.2mm}{
    \begin{tabular}{c|ccccc}
    \hline \hline
    Datasets&\#Nodes&\#Edges&\#Features&\#Classes&\#Average Degree\\
    \hline
         Cora~\cite{mccallum2000automating}&2708&5429&1433&7&2.00\\
         Cora\_ML~\cite{mccallum2000automating}&2810&7981&2879&7&2.84\\
         Citeseer~\cite{mccallum2000automating}&3327&4732&3703&6&1.42\\
         Pol.Blogs~\cite{adamic2005political}&1222&16714&/&2&13.68\\
         Pubmed~\cite{sen2008collective}&19717&44325&500&3&2.25\\
    \hline \hline
    \end{tabular}}
\end{table}

\subsection{Local GNN Model}
To demonstrate that Graph-Fraudster is effective in various GNN structures based GVFLs, three GNN models are adopted as local participants.
\begin{itemize}
    \item \textbf{Graph convolutional network (GCN)~\cite{kipf2016semi}:} it uses a layer-wise propagation rule based on a fist-order approximation of spectral convolutions on graphs. For node classification, a two-layer GCN is adopted as each local GNN model in GVFL. The node is represented as:
    \begin{equation}
        h_{i}^{(l+1)}=\rho\left(\sum_{j \in ne(i)} \frac{1}{\sqrt{\tilde{D}_{ii} \tilde{D}_{jj}}} h_{j}^{(l)} W^{(l)}\right)
    \end{equation}
    where $h^{(l+1)}$ is $(l+1)$-th layer's node representation and $W^{(l)}$ is the parameters of $l$-th layer. Node $j$ belongs to neighbor node set $ne$ of node $i$. $\Tilde{D}_{ii}=\sum_{j}(A+I_{N})$ is the degree matrix of $A+I_{N}$, and $I_N$ is the identity matrix.
    \item \textbf{Simple graph convolution (SGC)~\cite{wu2019simplifying}:} it is a linearized version of GCN without the activation function. The aggregation function can be formulated as:
    \begin{equation}
        h_{i}^{(l+1)}=\sum_{j \in n e(i)} \frac{1}{\sqrt{\tilde{D}_{ii} \tilde{D}_{jj}}} h_{j}^{(l)} W^{(l)}
    \end{equation}
    SGC usually outperforms GCN for node classification.
    \item \textbf{Robust GCN (RGCN)~\cite{zhu2019robust}:} it adopts Gaussian distributions as the hidden representations of nodes, which can absorb the effects of adversarial attack:
    \begin{equation}
    \footnotesize
    \begin{split}
        &\mu_{i}^{(l+1)}=\rho\left(\sum_{j \in ne(i)} \frac{1}{\sqrt{\tilde{D}_{ii} \tilde{D}_{jj}}}\left(\mu_{j}^{(l)} \odot \alpha_{j}^{(l)}\right) W_{\mu}^{(l)}\right)\\
    &\sigma_{i}^{(l+1)}=\rho\left(\sum_{j \in n e(i)} \frac{1}{\tilde{D}_{ii} \tilde{D}_{jj}}\left(\sigma_{j}^{(l)} \odot \alpha_{j}^{(l)} \odot \alpha_{j}^{(l)}\right) W_{\sigma}^{(l)}\right)
    \end{split}
    \end{equation}
    where $\mu$ is the means and $\sigma$ is the variances of Gaussian distributions. $W_{\mu}$ and $W_{\sigma}$ are weight matrix of the means and the variances, respectively. $\odot$ is the element-wise product.
\end{itemize}

\begin{table*}
\caption{The accuracy(±STD) of GVFL based on different GNN models against four adversarial attack methods on multiple datasets (in dual-participants case).}
\label{MainTask}
\centering
\small
\renewcommand\arraystretch{1.3}
\resizebox{\linewidth}{!}{
\begin{tabular}{c|c|c|ccccc}
\hline
\hline
\multirow{2}{*}{\textbf{Local Model}} & \multirow{2}{*}{\textbf{Datasets}} &  \multirow{2}{*}{\textbf{Centralized}} &\multicolumn{5}{c}{\textbf{Method}}                                                    \\ \cline{4-8}
& & &\textbf{Clean} & \textbf{RND} & \textbf{NETTACK} & \textbf{FGA} & \textbf{Graph-Fraudster}   \\
\hline
\multirow{5}{*}{\textbf{GCN}}   
& \textbf{Cora} &  0.806  & 0.719±0.017    & 0.663±0.015  & 0.541±0.043      & 0.565±0.051  & \textbf{0.435±0.034} \\
& \textbf{Cora\_ML}& 0.843 & 0.809±0.008    & 0.766±0.011  & 0.600±0.046      & 0.654±0.035  & \textbf{0.480±0.032} \\
& \textbf{Citeseer}& 0.682 & 0.629±0.019    & 0.562±0.033  & 0.507±0.032      & 0.496±0.037  & \textbf{0.376±0.028} \\
& \textbf{Pol.Blogs}& 0.955 & 0.923±0.015    & 0.835±0.024  & 0.775±0.058      & 0.745±0.024  & \textbf{0.713±0.032} \\
& \textbf{Pubmed}&  0.789 & 0.718±0.022    & 0.645±0.011  & 0.381±0.042      & 0.418±0.053  & \textbf{0.344±0.031} \\
\hline
\multirow{5}{*}{\textbf{SGC}}   
& \textbf{Cora}&  0.808   & 0.722±0.015    & 0.669±0.020  & 0.544±0.052      & 0.564±0.062  & \textbf{0.448±0.033} \\
& \textbf{Cora\_ML}& 0.848& 0.814±0.011    & 0.768±0.013  & 0.593±0.049      & 0.625±0.034  & \textbf{0.491±0.018} \\
& \textbf{Citeseer}&0.689 & 0.637±0.021    & 0.565±0.032  & 0.507±0.028      & 0.483±0.049  & \textbf{0.367±0.019} \\
& \textbf{Pol.Blogs}& 0.954& 0.922±0.015    & 0.844±0.027  & 0.783±0.053      & 0.770±0.039  & \textbf{0.714±0.036} \\
& \textbf{Pubmed}& 0.788  & 0.718±0.019    & 0.647±0.012  & 0.381±0.033      & 0.412±0.055  & \textbf{0.339±0.026} \\
\hline
\multirow{5}{*}{\textbf{RGCN}}  
& \textbf{Cora}&  0.808   & 0.725±0.009    & 0.671±0.012  & 0.518±0.026      & 0.581±0.036  & \textbf{0.454±0.027} \\
& \textbf{Cora\_ML}& 0.845 & 0.812±0.010    & 0.769±0.016  & 0.620±0.071      & 0.681±0.068  & \textbf{0.529±0.042} \\
& \textbf{Citeseer}& 0.674 & 0.642±0.013    & 0.566±0.016  & 0.503±0.037      & 0.519±0.029  & \textbf{0.386±0.038} \\
& \textbf{Pol.Blogs}& 0.954 & 0.949±0.005    & 0.833±0.026  & 0.785±0.074      & 0.775±0.049  & \textbf{0.708±0.018} \\
& \textbf{Pubmed}& 0.789 & 0.740±0.013    & 0.662±0.018  & 0.387±0.037               & 0.431±0.044            & \textbf{0.352±0.017}                  \\
\hline
\hline
\end{tabular}}
\end{table*}

\subsection{Attack Baselines}
Since this is the first work of adversarial attack on GVFL, three attack methods in the centralized setting are chosen as baselines. To make the attack methods transferable to GVFL, the probabilities returned by the server model and the local data are used to train a surrogate model, which is the same as the local GNN model. The baselines are briefly described as follows.
\begin{itemize}
    \item \textbf{RND~\cite{zugner2018adversarial}:} we assume that the connection of different types of nodes will affect the prediction result. For a given target node, RND randomly samples a node whose predicted label is unequal to the target node. Then, add an edge between the target node and the sampled node. 
    \item \textbf{NETTACK~\cite{zugner2018adversarial}:} it generates the adversarial edges guided by two evaluation functions from selected candidate edges and features. It modifies the highest-scoring edges or features, and updates the adversarial network iteratively to confuse GNNs.
    \item \textbf{FGA~\cite{chen2018fast}:} it constructs the edge gradient network based on original network firstly. Then, it generates the adversarial edges iteratively with the maximal absolute edge gradient til it reaches the attack budget $\Delta$.
\end{itemize}

\subsection{Experiment Setup}
For each local GNN model, a two-layer GNN model is applied to extract the local node embeddings, whose dimension is set to 16. The number of hidden units is fixed to 32. For GCN and SGC, the activation function is ReLU. ELU and ReLU are applied for means and variances respectively in RGCN. The GVFL is trained for 200 epochs using Adam with learning rate of 0.01. Considering sparsity of the dataset and concealment of attack, the attack budget $\Delta$ is fixed to 1. Besides, accuracy of the server model is applied as the main evaluation metric, and the lower the accuracy, the better the attacker's performance.

Our experimental environment consists of Intel XEON 6240 2.6GHz x 18C (CPU), Tesla V100 32GiB (GPU), 16GiB memory (DDR4-RECC 2666) and Ubuntu 16.04 (OS).

\subsection{Attack Performance}
In this subsection, Graph-Fraudster is conducted on five real-world datasets in two main scenes, i.e., dual-participants based GVFL and multi-participant based GVFL.
\subsubsection{Attack on Dual-participants Based GVFL} As described above, in this setting, datasets are divided into two pieces both in edges and node features randomly. Multi-perspective metrics are used to evaluate the performance of the attackers, including the node classification accuracy, precision, recall, F1-Score, mean absolute error (MAE) and Log Loss.

\begin{figure*}
  \centering 
\includegraphics[width=7 in]{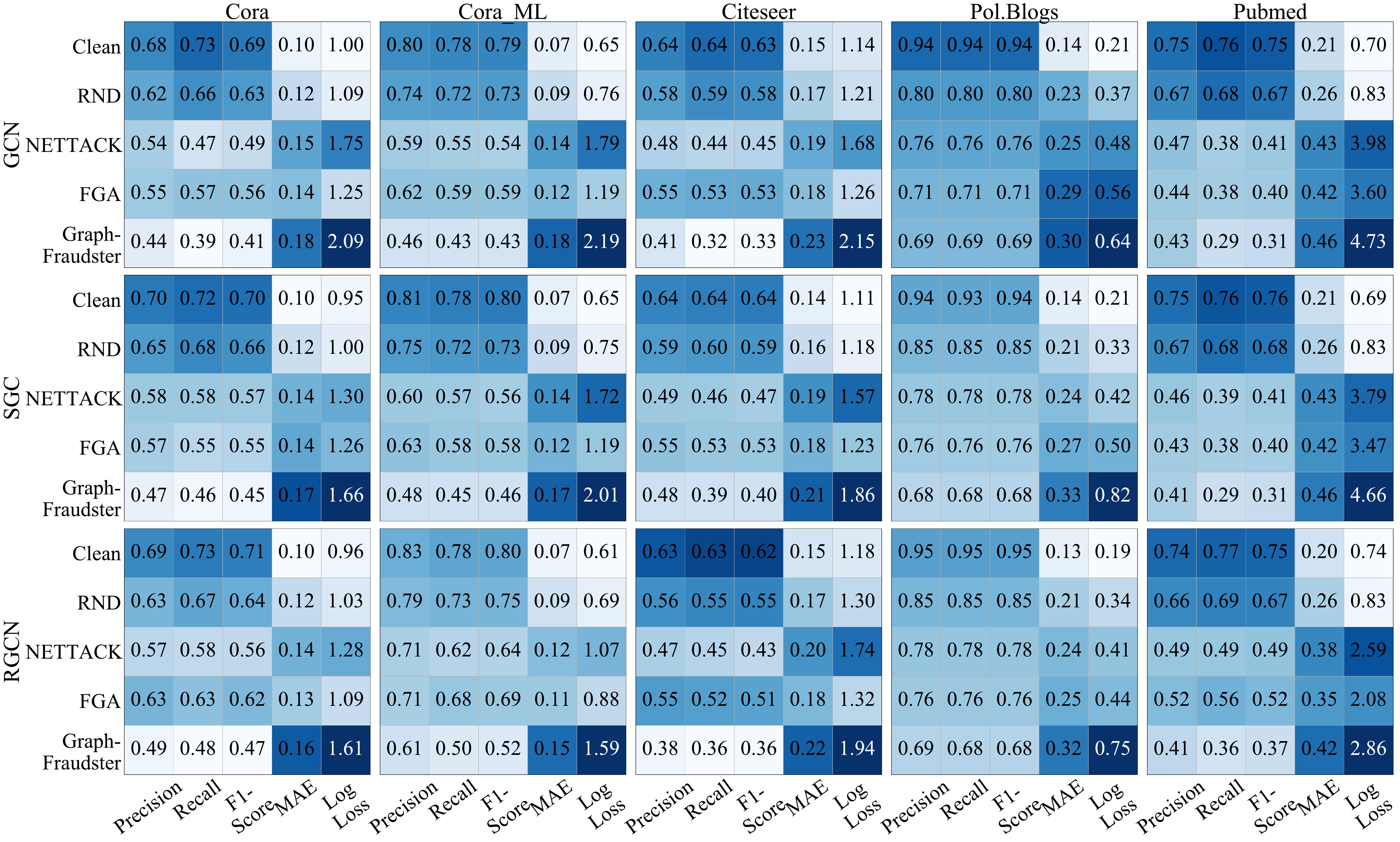}
  \caption{Multi-perspective metrics to measure the performance of attacks. For precision, recall, and F1-Score, the lower the attacker’s score, the better the attack performance is, and for MAE and Log Loss, the higher the attacker’s score, the better the attack performance is.}
  \label{fig:metrics} 
\end{figure*}

As shown in TABLE~\ref{MainTask}, in order to verify the generality of Graph-Fraudster in different graph segmentation situations, we conduct the experiments 10 times and report the average accuracy with standard deviation. The best attack performance is highlighted in bold.  In Fig. \ref{fig:metrics}, other five metrics are shown to evaluate the proposed method variously. Specifically, precision, recall and F1-Score are applied to measure the performance of GVFL. The lower the attacker's score, the better the attack performance is. MAE and Log Loss are used to evaluate the degree of confusion of the target server model. Higher value represents better performance of attack. Some observations are concluded in this experiment.

\begin{figure*}
  \centering 
\includegraphics[width=7 in]{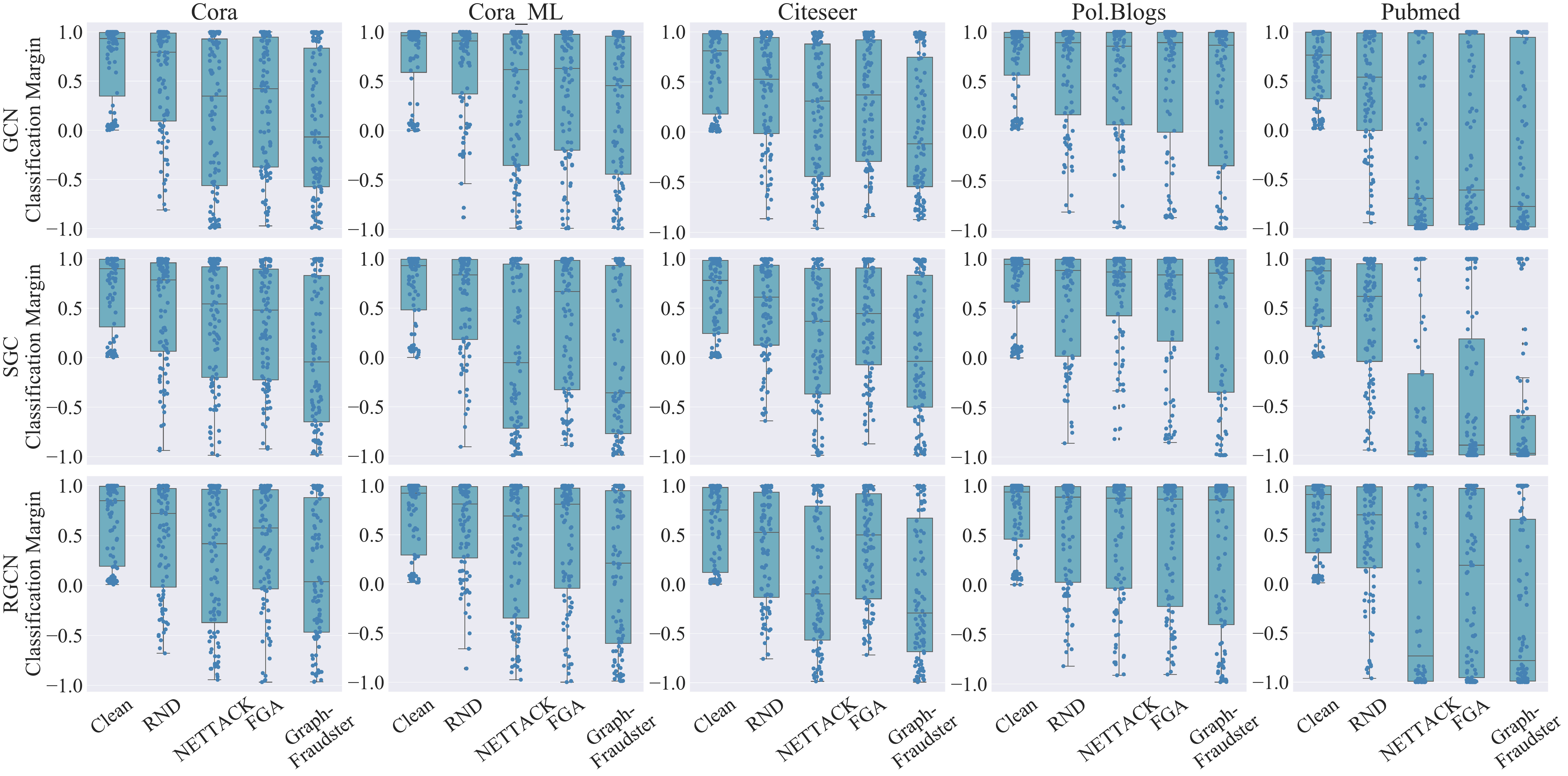}
  \caption{Classification margin of target nodes predicted by GVFL on five datasets. The lower classification margin, the better attack performance is.}
  \label{fig:CM} 
\end{figure*}

\begin{figure*}
  \centering 
\includegraphics[width=7 in]{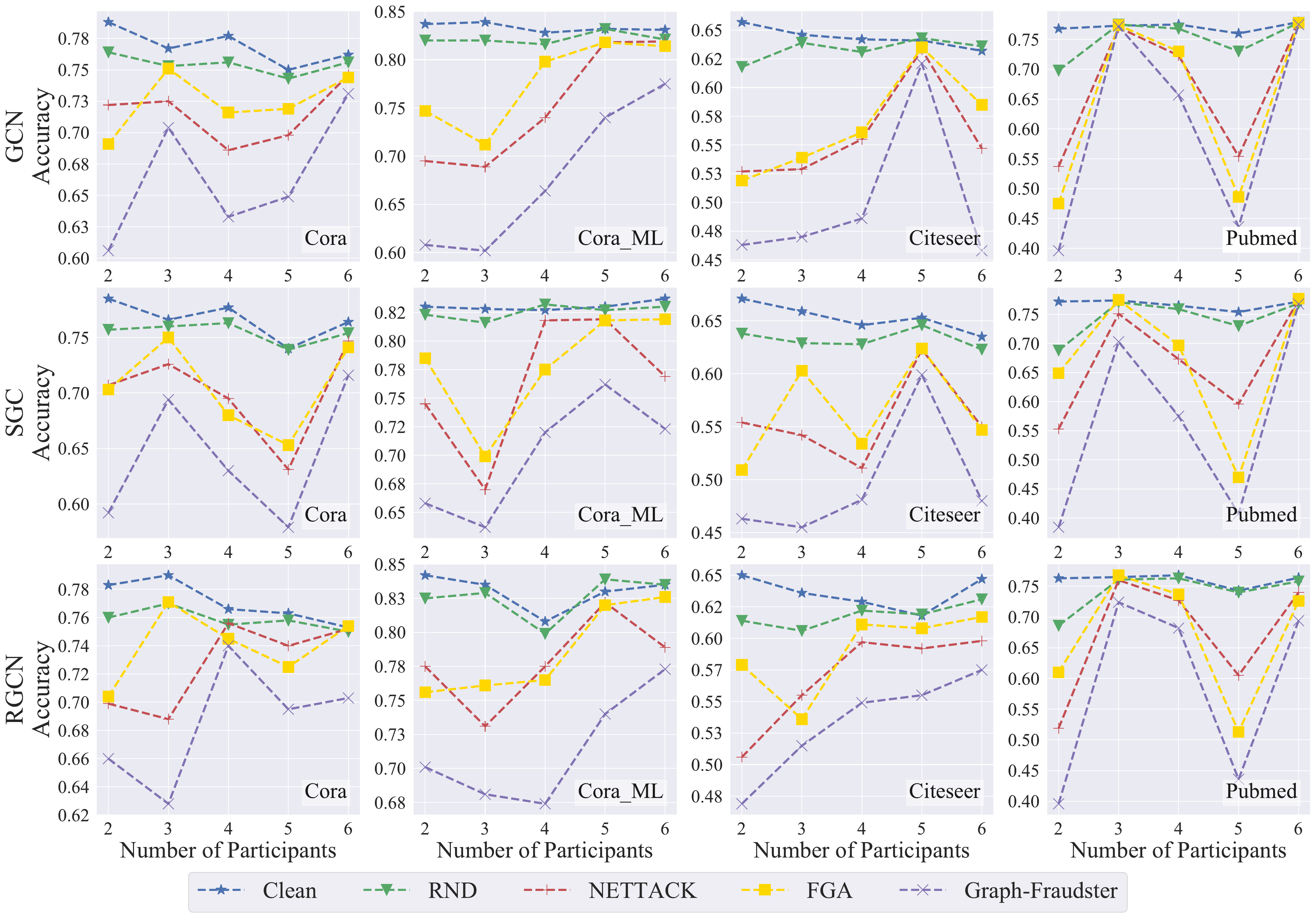}
  \caption{Attack on multi-participant based GVFL. The x-axis represents the number of participants, and the y-axis represents node classification tasks' accuracy. Each row represents experiments on different datasets of the same local GNN model.}
  \label{fig:multi} 
\end{figure*}

\begin{itemize}
    \item \emph{Graph-Fraudster achieves the SOTA attack performance compared with baselines.} For instance, the  accuracy of Graph-Fraudster degrades the baselines more than 10\% on Cora, Cora\_ML and Citeseer, which shows the superior performance of Graph-Fraudster. For Pol.Blogs, it can be seen that the performance of GVFL drops less than other datasets under attacks. For example, Graph-Fraudster reduces the GVFL's performance by about 21\%, 20.8\%, 24.1\%, corresponding to GCN, SGC and RGCN respectively, but gains more than 25\% on the other datasets. It is caused by the rich relationships of Pol.Blogs which make attacks more difficult within limited perturbation budget. It’s worth mentioning that the standard deviation shows that Graph-Fraudster is more stable than NETTACK and FGA generally. Fig. \ref{fig:metrics} reports the results on other five metrics, and Graph-Fraudster gains better attack performance than baselines on each metric.
    \item \emph{The integrity of graph-structured data affects the performance of the model.} Compared with the performance of centralized GNN models, GVFL's performance drops. It can be analyzed that the local GNN models cannot capture complete node information while edges are assigned to different participants, which weakens the quality of the local node embeddings. Furthermore,  GVFL performs better on the graph with higher degree, e.g., the performance of GVFL based on GCN drops by 8.7\% on Cora (the average degree is 2.00) while it only drops by 3.2\% on Pol.Blogs (the average degree is 13.68). There are similar results on different GVFLs. The phenomenon can be interpreted as the denser the graph can retain more relational information after data segmentation.
    \item \emph{Graph-Fraudster outperforms the other baselines on the robust GNN model.} RGCN was proposed as a kind of robust GNN model with defense mechanism. Thus, in most cases, RGCN-based GVFL outperforms other GVFLs without defense mechanism. Despite the performance of attack methods declines, Graph-Fraudster still keeps the state-of-the-art performance. Additionally, benefiting from its randomness, RND always maintains its performance, but its attack performance is poor.
\end{itemize}

\begin{figure*}
  \centering 
    \includegraphics[width=5.8in]{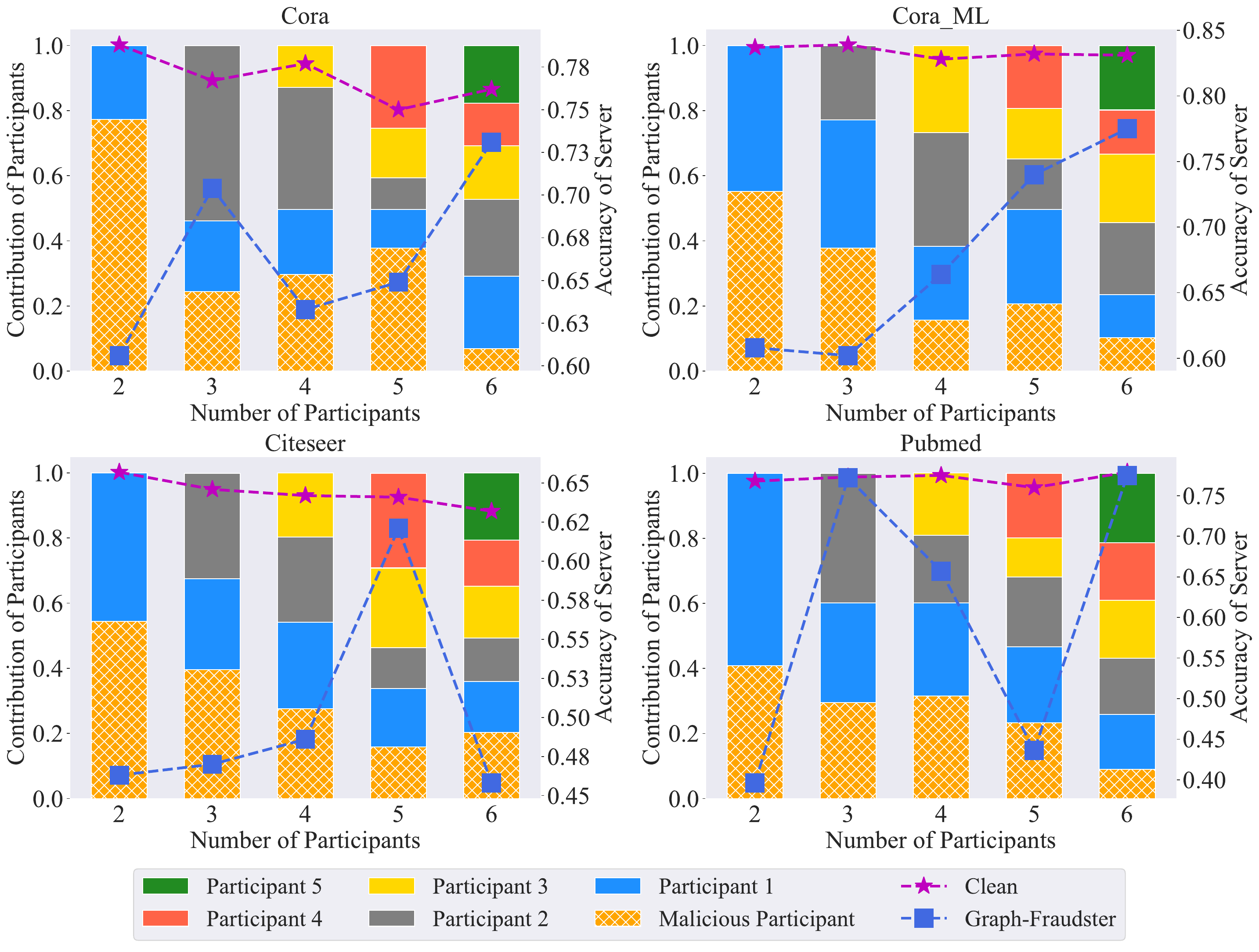}
  \caption{Relationship between attack performance of Graph-Fraudster and contribution of the malicious participant on GCN-based VFL.}
  \label{fig:contribution} 
\end{figure*}

To further compare the performance of attack methods in more details, classification margin is used as a metric to measure the influence of the attack methods on the probabilities of the model's prediction, following Zügner's work~\cite{zugner2018adversarial}. We select 100 nodes from the test set, which are correctly classified. These nodes satisfy:
\begin{itemize}
    \item the 20 nodes with the highest classification margin.
    \item the 20 nodes with the lowest classification margin but still be classified correctly.
    \item the 60 nodes are selected randomly.
\end{itemize}

It can be observed from Fig. \ref{fig:CM} that Graph-Fraudster outperforms baselines. More than half of the target nodes' classification margin below 0, on Cora, Citeseer and Pubmed, corresponding to the higher attack performance of Graph-Fraudster on these datasets. Moreover, more nodes gain low classification margin than NETTACK and FGA, under Graph-Fraudster's attack, on Cora\_ML. Although, for Pol.Blogs, all attack methods cannot achieve the same performance as well as other datasets due to its density. Graph-Fraudster still performs better. On Pubmed, Graph-Fraudster achieves similar performance to NETTACK and FGA, but stays ahead. For RND, only a few nodes obtains low classification margin, which explains its worst attack performance. Additionally, there are still some nodes with higher classification margin, corresponding to nodes that have not been successfully attacked. It may be caused by the single-edge attack, which is unable to make the GVFL misclassify these nodes.

The above experiments suggest that GVFL suffers from adversarial attacks. Although the existing adversarial attacks transferable to GVFL are not powerful enough without the knowledge of GVFL, GVFL still performs as expected. It reveals that GVFL is vulnerable to adversarial attacks. The vulnerability may be caused by the GNN model. The adversarial attacks make the GNN model extract the node embedding vector incorrectly and finally confuse the server model. Further, benefiting from the global node embeddings leakage in GVFL, Graph-Fraudster establishes a shadow server model which guides the attack to succeed. More knowledge of GVFL helps the performance of attacks, which explains the superior performance of Graph-Fraudster.

\begin{figure*}
	\centering  
	\vspace{-0.35cm} 
	\subfigtopskip=2pt 
	\subfigbottomskip=2pt 
	\subfigcapskip=-5pt
	\subfigure[GCN (Clean)]{
		\label{tsne.sub.1}
		\includegraphics[width=0.26\linewidth]{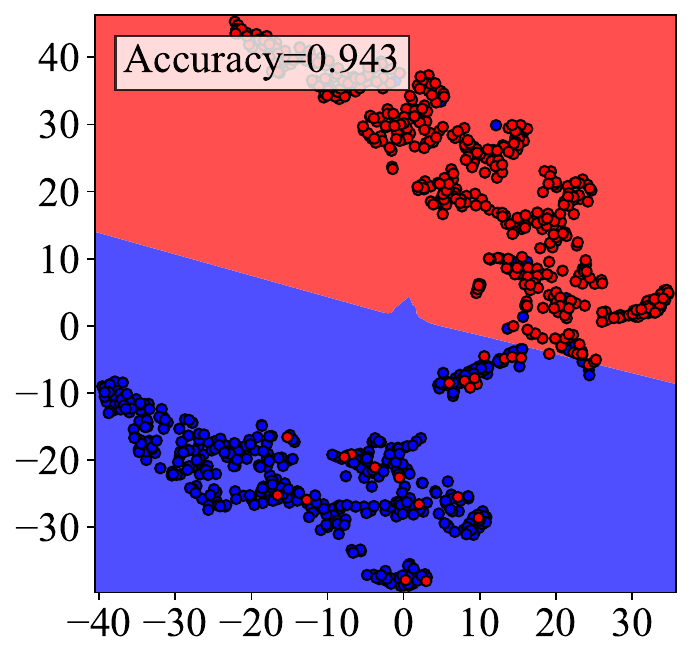}}
	\subfigure[SGC (Clean)]{
		\label{tsne.sub.2}
		\includegraphics[width=0.26\linewidth]{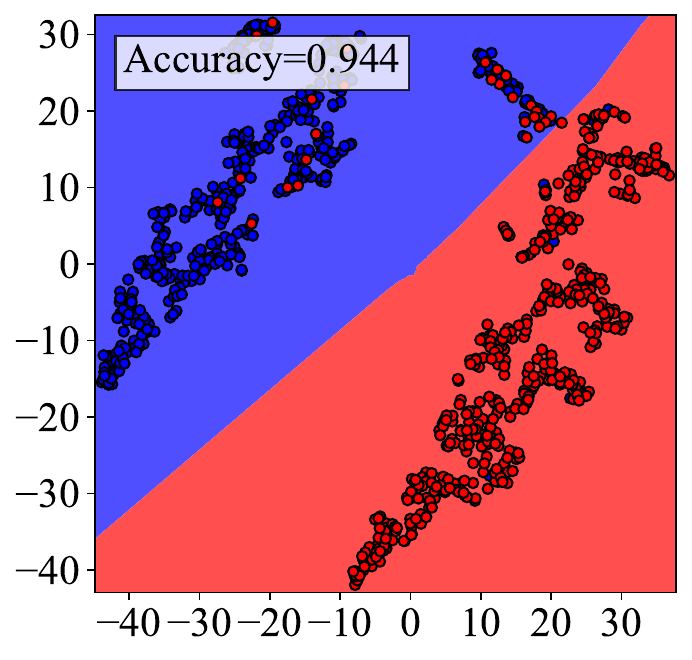}}
	\subfigure[RGCN (Clean)]{
		\label{tsne.sub.3}
		\includegraphics[width=0.26\linewidth]{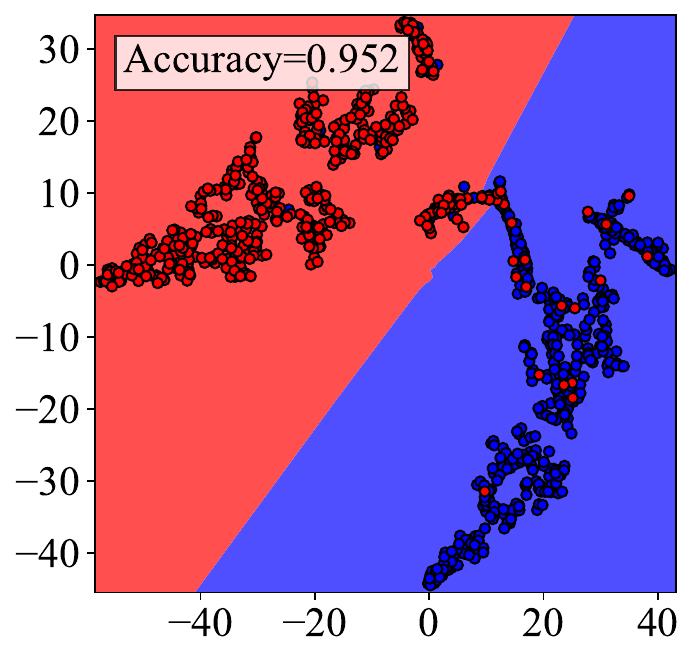}}
	
	\subfigure[GCN (Graph-Fraudster)]{
		\label{tsne.sub.4}
		\includegraphics[width=0.26\linewidth]{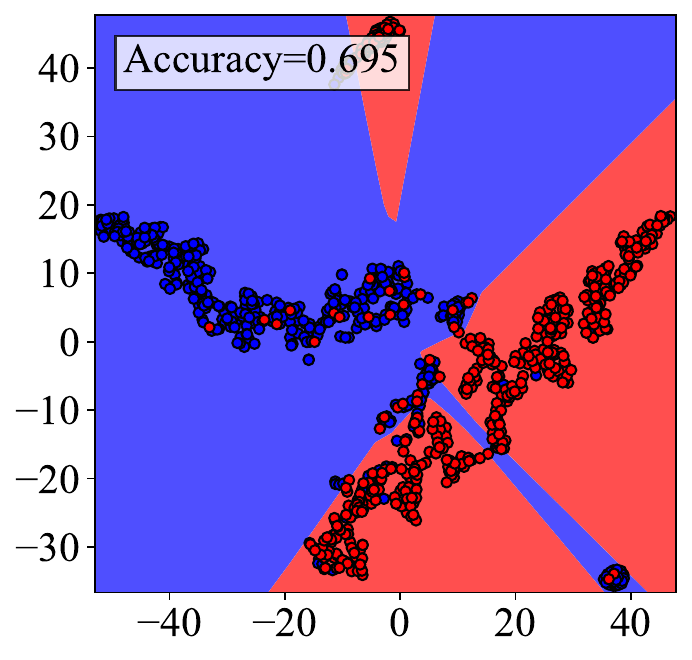}}
	\subfigure[SGC (Graph-Fraudster)]{
		\label{tsne.sub.5}
		\includegraphics[width=0.26\linewidth]{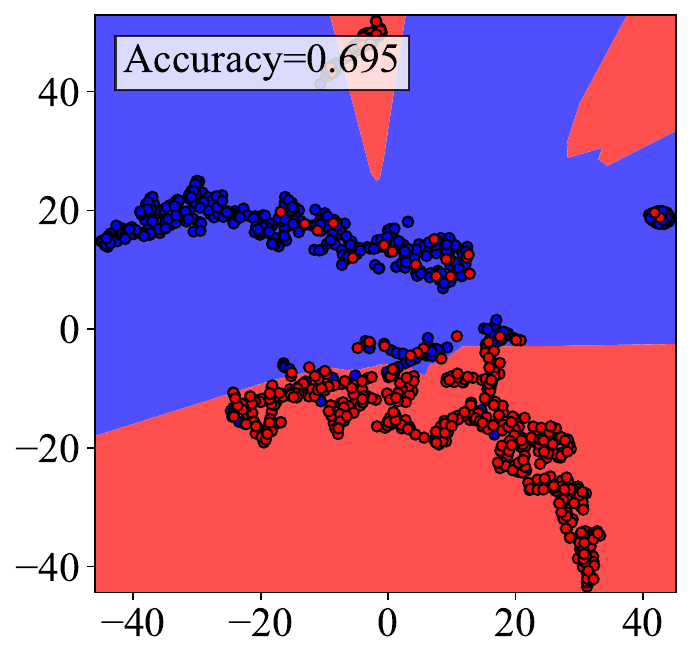}}
	\subfigure[RGCN (Graph-Fraudster)]{
		\label{tsne.sub.6}
		\includegraphics[width=0.26\linewidth]{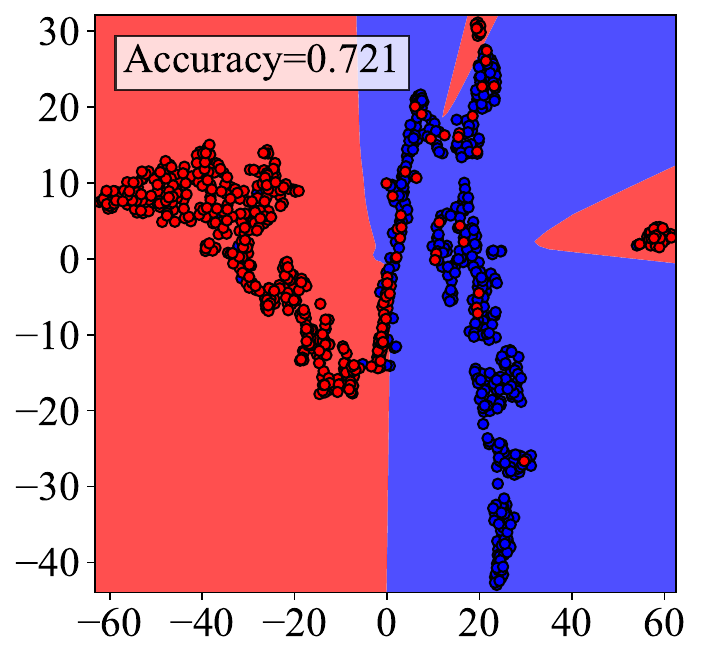}}
	\caption{Decision boundary before (1st row) and after (2nd row) Graph-Fraudster's attack on Pol.Blogs.}
	\label{fig:tsne}
\end{figure*}

\subsubsection{Attack on Multi-participant Based GVFL}
More generally, there are multiple participants in FL. In order to verify the capacity of Graph-Fraudster, the number of participants is set from 2 to 6 in the experiments. To avoid the emergence of numerous isolated nodes, we just distribute node features randomly to each participant without segmentation of edges. Since Pol.Blogs has no node features, the experiments will be conducted on the remaining four datasets. The results are shown in Fig. \ref{fig:multi}. 

First, comparing to edges segmentation, the performance of GVFL is more close to the centralized GNN models in this setting due to the retention of complete relationships between node pairs. Then, as the number of participants increases, the accuracy of GVFL decreases gradually. It is caused by information loss in node features segmentation. Same as edges segmentation setting, Graph-Fraudster maintains optimal attack performance with two participants. In general, the performance of attack methods decreases with the increasing of participants number. Because the server model treats every participant equally, as the number of participants increases, the contribution of the malicious participant's node embeddings decreases. As a result, the server model is harder to be confused. For example, on Pubmed, all attack methods almost fail with the single-edge attack when the participant number is 6. Interestingly, it not always happens in the experiments. Taking Citeseer as an example, the accuracy of GVFL falls when the number of participants are 6. We give the relationship between attack performance of Graph-Fraudster and contribution of the malicious participant in Fig. \ref{fig:contribution}. Contribution of $i$-th participant is defined as $C_{i}=\frac{acc_{i}}{\sum^{K}_{j}{acc_{j=1}}}$, where $K$ is the number of participants and $acc_{i}$ is the accuracy of the server model when only participant $i$'s node embeddings is input. It can be observed that on Cora\_ML, when the number of participants increases, the contribution of participants gradually decreases, and the server model is less susceptible to attacks. On Citeseer, when the number of participants is 6, the proportion of malicious participant's contribution increases, causing the server model to be attacked by Graph-Fraudster. We attribute the reason to the segmentation of the dataset. In this setting, the node features are divided randomly, which may cause that not all participants get useful node features. Thus, a small perturbation is powerful enough to fool the server model. On the contrary, the performance drops when the benign participants get some key node features and the slight structural perturbation from malicious participants is not enough to affect the performance of the server model. That is, the model is biased against the data. Besides, the properties of datasets affect attack performance. Taking Citeseer, which is the most sparse, and it can be seen that Graph-Fraudster maintains good attack performance. Due to its sparseness, it improves the influence of structural attacks on the perturbations of the local node embeddings. 

In summary, in the multi-participant case, with the number of participants increases, the performance of attackers generally shows a downward trend. It may be caused by the influence of the server model of each participant decreases while the number of participants increases. Thus, considering adding perturbations into both structure and node features may a better strategy, and it will be considered in our future work. Further, the model's bias against the data is one of the possible reasons for the success of the adversarial attack. It indicates data bias of the model may lead the server model more rely on high-contributing participants. As a result, it is easier for attacks which initiated by high-contributing participants to cheat the server model. Therefore, setting up a fair evaluation mechanism for participants may help to improve the robustness of GVFL against adversarial attacks.

\subsubsection{The Impact of Attacks on Decision Boundary}
To explain the attack results of Graph-Fraudster better, the decision boundaries on Pol.Blogs before and after attack are plotted in Fig. \ref{fig:tsne}. The setting is the same as dual-participants based GVFL. Combining the accuracy of the server model, the decision boundary is regular without perturbations, which corresponds to high accuracy. After attack, as mentioned in the previous experiments, the server model is not easy to be perturbed with high accuracy. However, some nodes break away from their corresponding groups, which makes the decision boundary fracture. In other words, these outliers will be misclassified.

\begin{figure*}[htbp]
  \centering 
  	\vspace{-0.35cm} 
	\subfigtopskip=2pt 
	\subfigbottomskip=2pt 
	\subfigcapskip=-5pt
	\subfigure[DP]{
		\label{defense.sub.1}
		\includegraphics[width=0.47\linewidth]{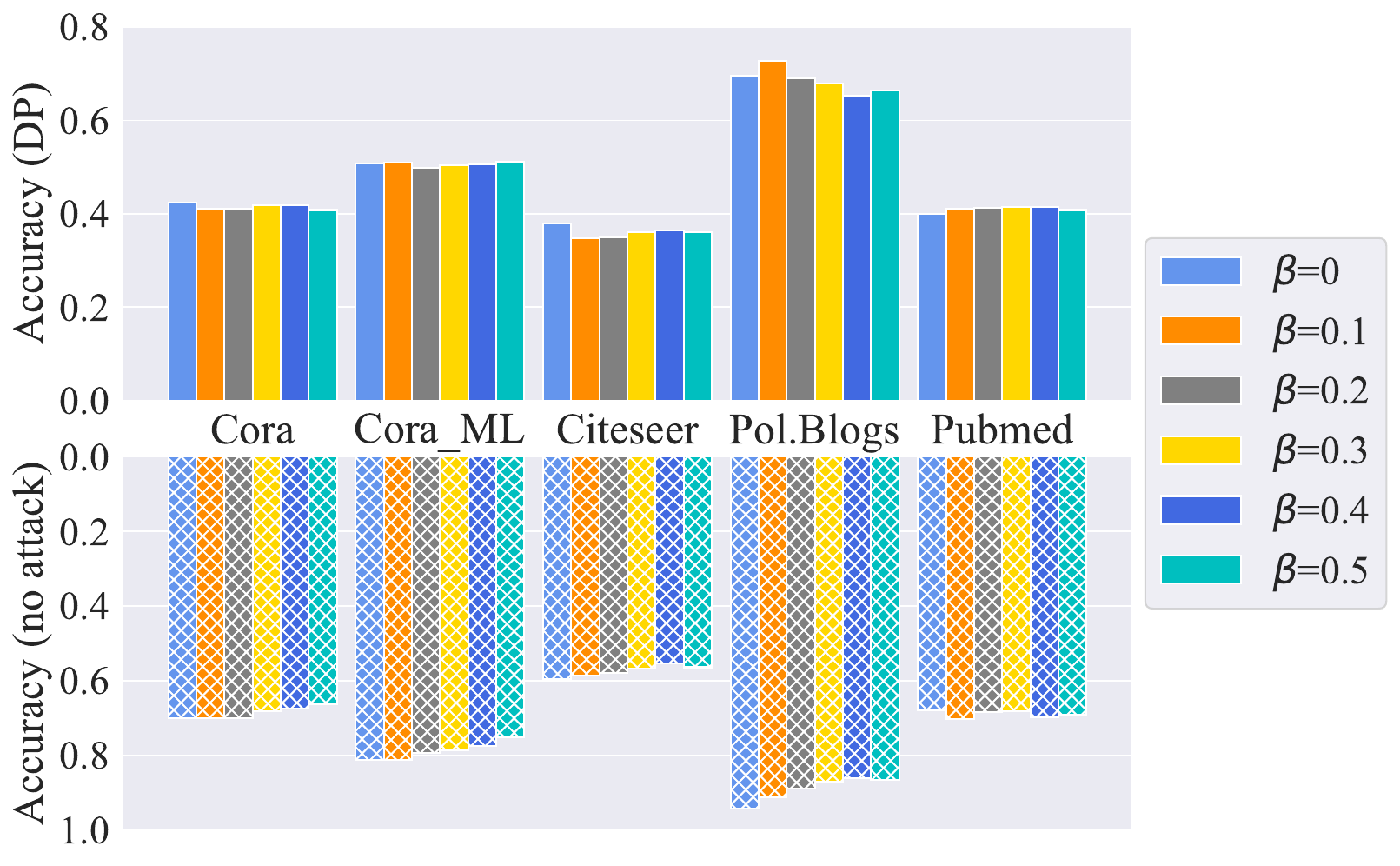}}
	\subfigure[Top-$k$]{
		\label{defense.sub.2}
		\includegraphics[width=0.475\linewidth]{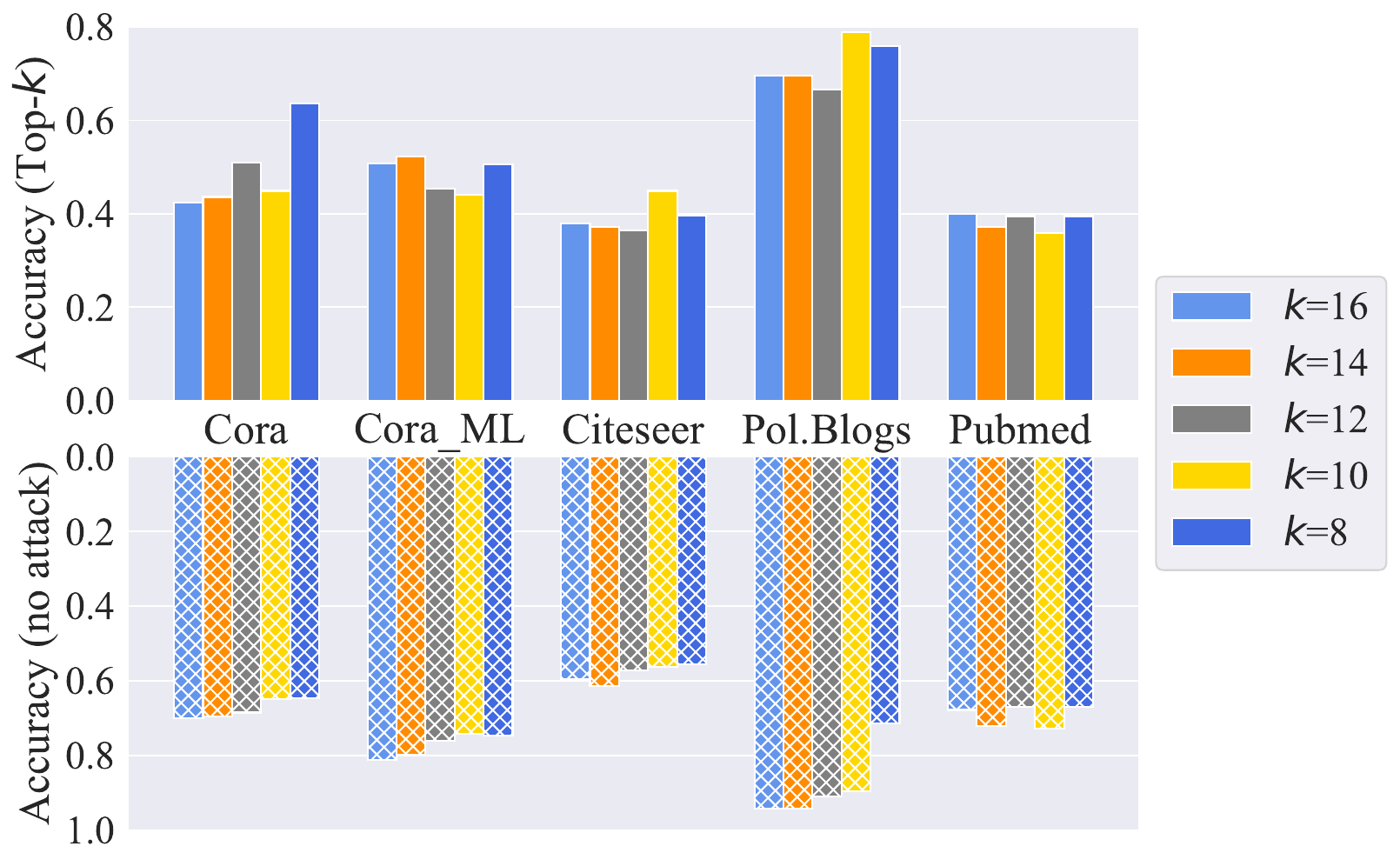}}
  \caption{Possible defense against Graph-Fraudster on GCN-based VFL. The upper part is the attack performance of Graph-Fraudster under the defense mechanisms, and the lower part is the performance of GVFL without attack.}
  \label{fig:defense} 
\end{figure*}

\begin{figure*}
	\centering  
	\vspace{-0.35cm} 
	\subfigtopskip=2pt 
	\subfigbottomskip=2pt 
	\subfigcapskip=-5pt
	\subfigure[GCN]{
		\label{dim.sub.1}
		\includegraphics[width=0.32\linewidth]{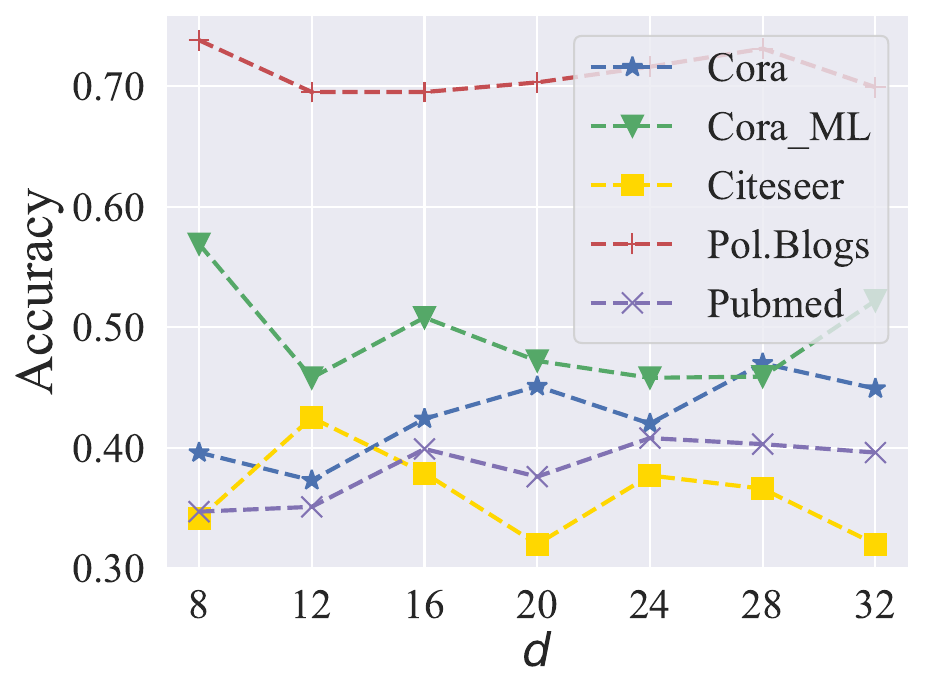}}
	\subfigure[SGC]{
		\label{dim.sub.2}
		\includegraphics[width=0.32\linewidth]{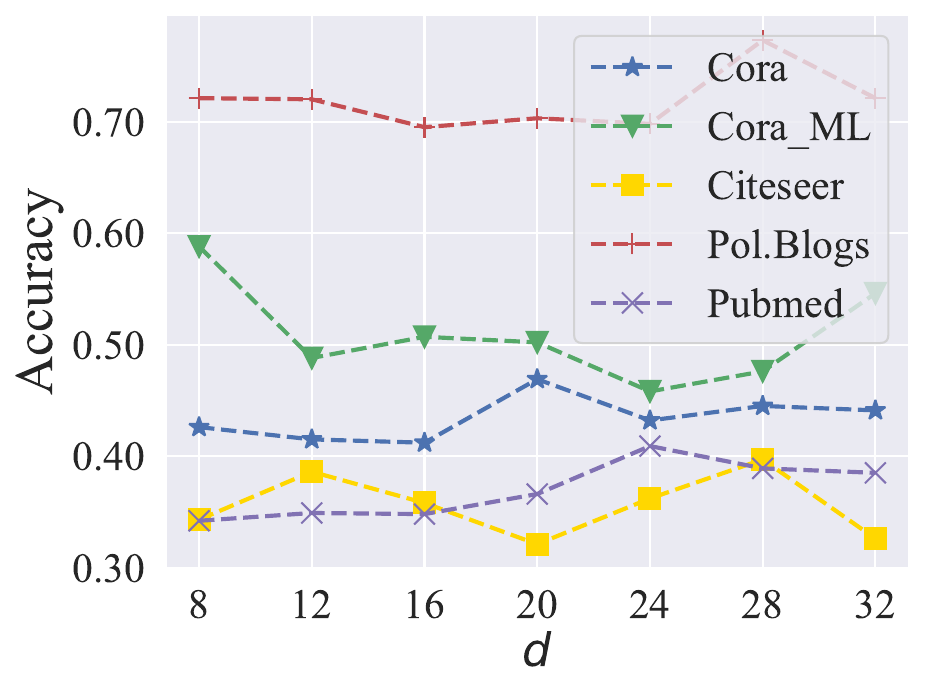}}
	\subfigure[RGCN]{
		\label{dim.sub.3}
		\includegraphics[width=0.32\linewidth]{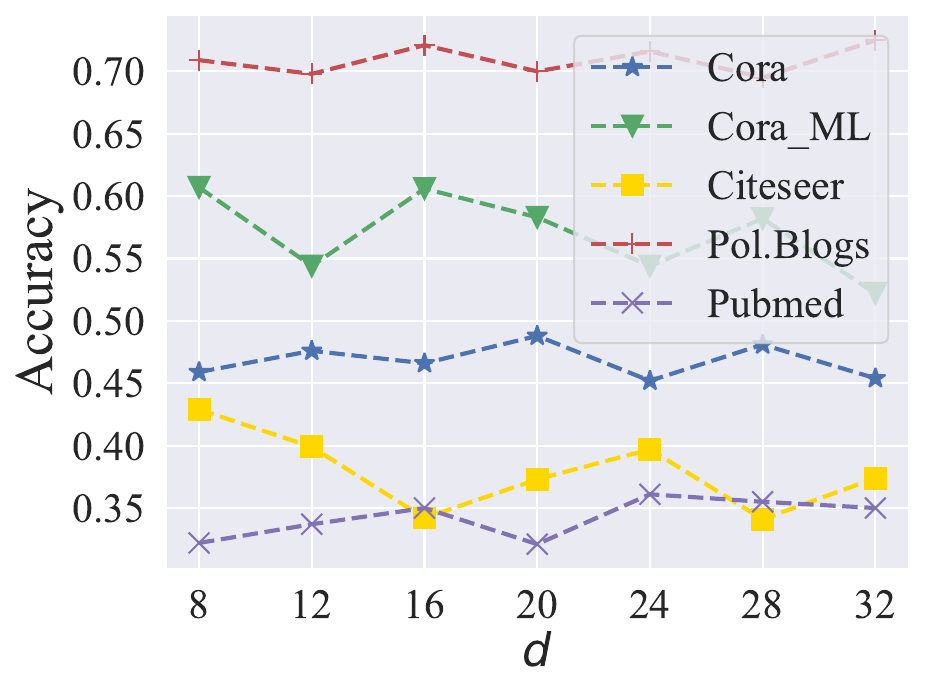}}

	\caption{Performance of Graph-Fraudster with different dimensions of the local node embeddings.}
	\label{fig:dim}
\end{figure*}

\subsection{Possible Defense}
To mitigate the performance of Graph-Fraudster, Differential Privacy (DP) mechanism and Top-$k$ mechanism are introduced as the possible defenses. DP is used to prevent Graph-Fraudster from stealing the node embeddings, and Top-$k$ devotes to filtering out perturbations in the local node embeddings. The two strategies are briefly introduced as follows:
\begin{itemize}
    \item \textbf{DP~\cite{dwork2008differential}:} randomized noise is injected into the local node embeddings before uploading, obeying the principle of DP. The Laplacian mechanism is used in this paper with the scale of noise $\beta$.
    \item \textbf{Top-$\bm{k}$~\cite{liu2020fedsel}:} before uploading the local node embeddings, we sort the embedding value and preserve the top $k$ values to reduce the influence of some useless information.
\end{itemize}

The scale of Laplacian noise $\beta$ is set from 0 to 0.5, and the value of $k$ is set from 8 to 16. Fig. \ref{fig:defense} shows the result on GCN-based GVFL. Note that $\beta=0$ and $k=16$ present the GVFL without defense mechanism. With the increase of $\beta$ or the decrease of $k$, these two defense mechanisms reduce the performance of GVFL to a certain extent. It shows that there should be a trade-off between defense performance and GVFL's performance. As shown in Fig. \ref{defense.sub.1}, DP has limited defensive capability against Graph-Fraudster, or even useless, e.g., the performance on Citeseer. It may be that the noise added by DP cannot offset the noise added by Graph-Fraudster well. As for Top-$k$, as shown in Fig. \ref{defense.sub.2}, it is more effective than DP, but it is unstable. For example, on Cora, Top-$k$ with $k=8$ outperforms other value and $k=12$ is second, which is not monotonous with the change of $k$. It is caused by the non-concentrated distribution of perturbations in the node embeddings. Specifically, the perturbations do not only exist in lower values. Moreover, the price of better defensive capability against Graph-Fraudster is the performance degradation of the main task. In summary, Graph-Fraudster can still perform well under two possible defense mechanisms. In future works, filtering perturbations from the local node embeddings is a feasible solution against adversarial attacks in GVFL, and keeping the performance of GVFL while gains defensive capability is also a great challenge.

\subsection{Parameter Analysis}
In this subsection, the sensitivity of the dimensions of the local node embeddings $d$ and the scale of noise $\epsilon$ for Graph-Fraudster is explored. The value of these parameters is adjusted to see how they affect the performance of Graph-Fraudster. In more detail, $d$ is set from 8 to 32 with a step size of 4, and $\epsilon$ is changed from 0 to 0.01.

\begin{figure}
  \centering 
  \includegraphics[width=2.5 in]{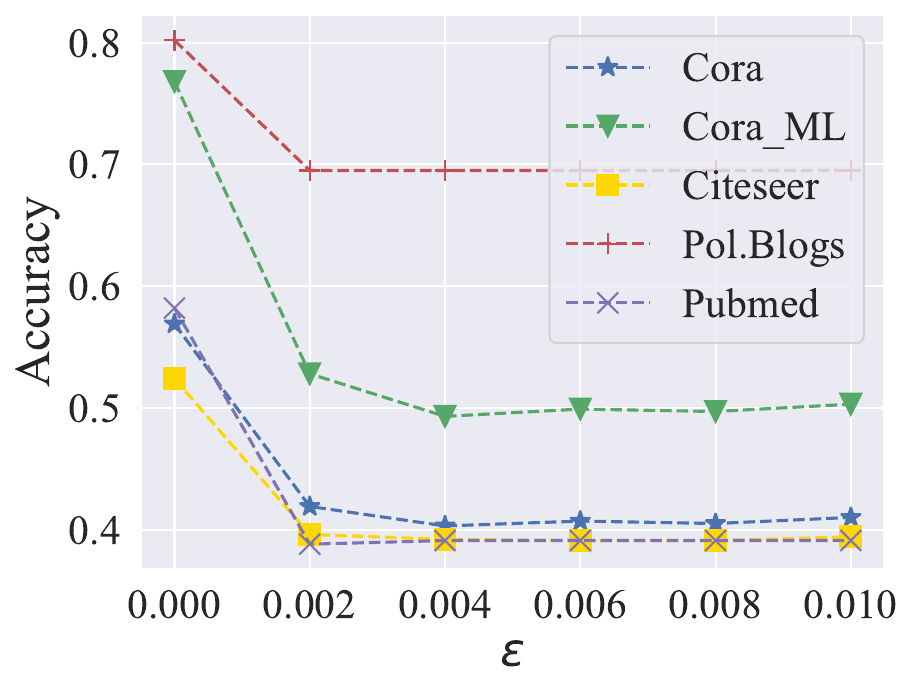}
  \caption{Performance of Graph-Fraudster with different noise scale $\epsilon$ on GCN-based VFL.}
  \label{fig:epsilon} 
\end{figure}

The performance change of Graph-Fraudster is shown in Fig. \ref{fig:dim} and Fig. \ref{fig:epsilon}. It can be observed that as the value of $d$ increases, the attack performance of Graph-Fraudster is not significantly affected (less than 10\%), which shows stability of Graph-Fraudster. In addition, we take the performance of Graph-Fraudster on GCN-based VFL with different noise scale $\epsilon$ as an example. The results are shown in Fig. \ref{fig:epsilon}. Similar results are observed in other settings. It can be seen that the accuracy of the server model drops rapidly in a small range of $\epsilon$ (from 0 to 0.002). Single-edge attack may be the main reason, because the perturbations are too small to further approximate the node embeddings with large-scale noise. Thus, for single-edge attack, a small value of $\epsilon$ is enough (e.g., 0.004 in this setting) to guide the attack, and choosing an appropriate value of $\epsilon$ can boost the performance of Graph-Fraudster.

\section{Conclusion}
\label{conclusion}
\underline{\textbf{\textit{Contributions.}}} The first novel adversarial attack method on GVFL, named Graph-Fraudster, is proposed by combining GVFL’s privacy leakage and the gradient of pairwise node. Benefiting from the information of GVFL, Graph-Fraudster achieves the state-of-the-art attack performance compared with the baselines. Extensive experiments have testified the effectiveness and efficiency of Graph-Fraudster and further revealed the vulnerability of GVFL even under defensive circumstances.

\underline{\textbf{\textit{Observations.}}} Besides of the adversarial attack problem formulation and Graph-Fraudster attack, several observations are gained according to the comprehensive experiments and analysis. (1) GVFL does suffer from adversarial attacks, especially when some information (e.g., node embeddings, structure of model and gradient of model etc.) is explored by the attackers; (2) graph splitting strategy of GVFL will not only affect the performance of GVFL, but also affect the success rate of adversarial attack, thus a proper splitting strategy should be seriously considered in real-world scenarios; (3) the more a participant contributes to the server, the more likely the perturbations added to the participant is to construct a successful adversarial attack; (4) although DP and Top-$K$ fail to fully defend Graph-Fraudster, the results show that denoising the uploaded node embeddings is still a feasible defensive strategy.

\underline{\textbf{\textit{Limitations.}}} Graph-Fraudster devotes to fooling the server model, even with single-edge modifying. However, the small number of perturbations does not mean that the perturbations are imperceptible, and these perturbations may change some properties of the graph significantly. Besides, frequent queries are needed for Graph-Fraudster, which are expensive and detectable.

\underline{\textbf{\textit{Future Works.}}} Adversarial attack on GVFL will raise our attention about the robustness of GVFL. There are two sides of suggestions are provided for future works. \textbf{Recommendations for vulnerability mining:} (1) For perturbation strategies, both structure and node features should be considered for perturbation, to reduce the impact of the data imbalance of attacks. (2) Since frequent queries on the server model are expensive and abnormal, reducing queries by stealing more information from GVFL is a possible solution. (3) Exploring the concealment of the perturbations for diverse downstream tasks, such as link prediction, graph classification, should be discussed as well. \textbf{Suggestions for defenders:} (1) Researchers are advised to focus on the information leakage problem on GVFL, which provides convenience for attackers. Preventing key information of GVFL from privacy inferencing is a direct and efficient method. (2) A fair evaluation agency for participants should be established. Over-reliance on high-contributing participants may bring potential security risks, i.e., the impact of adversarial attacks may be amplified by the data bias of the model. (3) Some defense or detection mechanisms can be deployed in terminals or servers to eliminate the impact of adversarial perturbations.


%





\ifCLASSOPTIONcaptionsoff
  \newpage
\fi



%
\bibliographystyle{IEEEtran}
\bibliography{ref}


%







\end{document}